\documentclass{article}

\usepackage[preprint]{neurips_2026}

\usepackage[utf8]{inputenc} 
\usepackage[T1]{fontenc}    
\usepackage{hyperref}       
\usepackage{url}            
\usepackage{booktabs}       
\usepackage{amsfonts}       
\usepackage{nicefrac}       
\usepackage{microtype}      
\usepackage{xcolor}         
\usepackage{enumitem}
\usepackage{amsmath}
\usepackage{amsthm}

\usepackage{titlesec}

\titlespacing*{\section}
  {0pt}        
  {6pt}        
  {3pt}        

\titlespacing*{\subsection}
  {0pt}        
  {4pt}        
  {2pt}        

\theoremstyle{plain}

\theoremstyle{definition}

\theoremstyle{remark}

\usepackage[textsize=tiny]{todonotes}

\usepackage{booktabs, multirow}
\usepackage{pifont} 
\renewcommand{\arraystretch}{1.2}  

\def\cY{\mathcal{Y}}

\def\R{\mathbb R}

\def\w {\mathbf{w}}

\def\x{\mathbf{x}}
\def\z{\mathbf{z}}

\def\y{\mathbf{y}}

\def\t{\mathbf{t}}

\title{ReSS: Learning Reasoning Models for Tabular Data Prediction via Symbolic Scaffold}

\author{%
  {\bf Chenlang Yi}$^{1}$\thanks{Equal contribution.} , 
  {\bf Gang Li}$^{1}\footnotemark[1]$ , 
  {\bf Zizhan Xiong}$^1$, 
  {\bf Tue Minh Cao}$^2$, \\
  {\bf Yanmin Gong}$^1$, 
  {\bf My T. Thai}$^2$, 
  {\bf Tianbao Yang}$^{1}$\thanks{Corresponding author. Email: \texttt{tianbao-yang@tamu.edu}} \\[1ex]
  $^1$Department of Computer Science \& Engineering, Texas A\&M University \\
  $^2$Department of Computer Science, University of Florida
}
\begin{document}

\maketitle

\begin{abstract}
Tabular data remains prevalent in high-stakes domains such as healthcare and finance, where predictive models are expected to provide both high accuracy and faithful, human-understandable reasoning. While symbolic models offer verifiable logic, they lack semantic expressiveness. Meanwhile, general-purpose LLMs often require specialized fine-tuning to master domain-specific tabular reasoning, which faces  dual challenges of scalable data curation and reasoning consistency. To address these challenges, we propose {\bf ReSS}, a systematic framework that bridges symbolic and neural reasoning models. ReSS leverages a decision-tree model to extract instance-level decision paths as symbolic scaffolds. These scaffolds, alongside input features and labels, guide an LLM to generate grounded natural-language reasoning that strictly adheres to the underlying decision logic. The resulting high-quality dataset is used to fine-tune a pretrained LLM into a specialized tabular reasoning model, further enhanced by a scaffold-invariant data augmentation strategy to improve generalization and explainability. To rigorously assess faithfulness, we introduce quantitative metrics including hallucination rate, explanation necessity, and explanation sufficiency. Experimental results on medical and financial benchmarks demonstrate that ReSS-trained models improve traditional decision trees and standard fine-tuning approaches up to $10\%$ while producing faithful and consistent reasoning.

\end{abstract}

 \setlength\abovedisplayskip{4pt}
\setlength\belowdisplayskip{4pt}
\setlength{\textfloatsep}{4pt}

\section{Introduction}
\label{sec:intro}
In high-stakes domains such as healthcare \cite{diabetes_130-us_hospitals_for_years_1999-2008_296}, finance \cite{bank_marketing_222} where tabular data is common, predictive accuracy alone is often insufficient. Models are also expected to provide faithful and human-understandable reasoning that enables practitioners to inspect, trust, and act upon their predictions. Despite decades of progress in tabular learning~\cite{DBLP:books/wa/BreimanFOS84,Chen_2016,si2024interpretabnet}, constructing models that jointly achieve strong performance and meaningful reasoning remains a fundamental challenge.

Traditional machine learning methods have long dominated tabular prediction, with tree-based approaches such as decision tree~\cite{DBLP:books/wa/BreimanFOS84}, random forests~\cite{10.1023/A:1010933404324}, XGBoost \cite{Chen_2016}, and LightGBM \cite{NIPS2017_6449f44a} achieving strong empirical performance. From a reasoning perspective, however, these models remain fundamentally limited. Tree ensemble methods do not expose an explicit, instance-level decision process, while single decision trees express reasoning solely through symbolic rules defined over low-level feature thresholds. Although such representations faithfully reflect the model’s internal logic, they lack semantic grounding in domain concepts and therefore cannot be readily interpreted by non-expert users, limiting their practical utility in real-world, high-stakes decision-making settings. Recent advances in deep learning (DL) for tabular data have produced models that are able to match or even surpass the performance of tree-based methods~\cite{huang2020tabtransformertabulardatamodeling,arik2020tabnetattentiveinterpretabletabular,abutbul2020dnfnetneuralarchitecturetabular,si2024interpretabnet}. However, these models usually provide limited or no interpretability regarding its decision process. Even when post-hoc explanations are available, they often fail to constitute truly human-understandable reasoning chains.


Recently, studies have leveraged LLMs for tabular data prediction~\cite{dinh2022liftlanguageinterfacedfinetuningnonlanguage, hegselmann2023tabllmfewshotclassificationtabular}. These methods transform tabular rows into natural language prompts to utilize the LLM’s prior knowledge for prediction. However, they essentially function as direct-prediction models rather than reasoning models, as they map inputs directly to final labels while bypassing the intermediate decision-making process. Consequently, they fail to provide a transparent, step-by-step logical chain to justify their predictions. 

This raises an important question: 
\textit{How to efficiently train a faithful reasoning model for tabular data prediction?}
There are two key challenges to be addressed: (i) how to curate high-quality reasoning data for fine-tuning an LLM; and (ii) how to ensure that the reasoning traces generated by the learned model are both faithful and explainable. Curating high-quality reasoning data for high-stakes domains such as healthcare and finance typically requires domain experts and is therefore expensive, time-consuming, and difficult to scale. Moreover, while current LLMs are capable of producing chain-of-thought (CoT)-like outputs, these explanations are not guaranteed to be faithful and causal to the tabular data prediction~\cite{https://doi.org/10.1111/nyas.15339,not_faithful_survey}.

To address these challenges, we propose {\bf ReSS}, a systematic framework that curates high-quality reasoning data via symbolic scaffolds and trains faithful reasoning models for tabular prediction. ReSS first trains a decision tree on tabular data and extracts instance-level decision paths from  samples that are correctly predicted by the decision tree, which serve as explicit symbolic scaffolds capturing the model’s underlying decision logic. These scaffolds, together with input features and labels, are used to guide an LLM to generate grounded, step-by-step natural-language reasoning that strictly adheres to the logical constraints encoded in the decision paths, yielding a reasoning dataset that is both faithful to the original decision process and interpretable by humans. 
To further enhance generalization and explainability, ReSS incorporates a scaffold-invariant data augmentation strategy that expands the reasoning dataset and introduces out-of-distribution examples by perturbing input features while preserving the symbolic decision paths and labels. 
We summarize our contributions as follows: 
\vspace*{-0.1in}
\begin{itemize}[leftmargin=*]
\item We propose ReSS, a systematic framework that leverages decision-tree paths as symbolic scaffolds to guide an LLM using its inherent domain knowledge to curate high-quality grounded reasoning data, and trains faithful reasoning models for tabular prediction.

\item We introduce a scaffold-invariant data augmentation strategy that preserves symbolic decision logic while generating both in-distribution and out-of-distribution reasoning data, enabling improved generalization and more robust explainability.

\item We propose a set of metrics to measure the faithfulness and explainability of generated CoT-like outputs by the trained reasoning model, including hallucination rate, explanation sufficiency, and explanation necessity.

\item We conduct experiments on five real-world datasets, including two from medicine and three from finance. The results demonstrate the effectiveness of ReSS and the scaffold-invariant data augmentation, and validate the faithfulness and causal impact of the generated rationales.
\end{itemize}



\section{Related work}

While tabular data prediction has long been dominated by tree-based ensemble models,  
recent years have seen the emergence of new approaches based on DL and LLMs.

{\bf Deep Learning Methods for Tabular Prediction.}
Early DL approaches for tabular prediction typically rely on generic architectures such as MLPs, which lack inductive biases aligned with the structure of tabular decision manifolds and often struggle to match the performance of tree-based methods~\cite{arik2020tabnetattentiveinterpretabletabular}. To overcome this limitation, subsequent work introduced tabular-specific inductive biases into neural architectures, including TabNet~\cite{arik2020tabnetattentiveinterpretabletabular}, DNF-Net~\cite{abutbul2020dnfnetneuralarchitecturetabular}, TabTransformer~\cite{huang2020tabtransformertabulardatamodeling},  TabPFN~\cite{hollmann2023tabpfntransformersolvessmall}.  But these methods are not interpretable. 
InterpreTabNet~\cite{si2024interpretabnet} attempts to provide intrinsic interpretability through instance-wise feature selection mechanisms, by
employing an LLM to map from the learned feature mask onto natural language text describing the learned signal.  Despite these advances, they do not aim to produce explicit, human-understandable reasoning processes.

{\bf LLMs for Tabular Prediction.} 
Recent work has explored applying LLMs to tabular prediction by serializing tabular records into natural-language descriptions, allowing LLMs to operate in their native textual space. Depending on whether model parameters are updated, these approaches can be broadly categorized into inference-only and fine-tuning–based methods. TABLET~\cite{slack2023tabletlearninginstructionstabular} studies inference-only prediction by prompting pretrained LLMs, but reports that simple tree-based baselines consistently outperform LLMs under comparable settings. Fine-tuning–based approaches, including LIFT~\cite{dinh2022liftlanguageinterfacedfinetuningnonlanguage} and TabLLM~\cite{hegselmann2023tabllmfewshotclassificationtabular}, demonstrate that language pretraining and meaningful feature semantics can improve tabular prediction, particularly in low-shot regimes. However, these gains are inconsistent across datasets, and overall performance still struggles to reliably surpass strong tree-based models. These existing methods focus solely on final predictions and do not generate explicit reasoning.

{\bf Large Reasoning Models.}
Recently, Large Reasoning Models (LRMs) trained by RL have demonstrated strong capabilities in solving complex tasks~\cite{guo2025deepseek,team2025kimi}, positioning them as a promising paradigm for tabular prediction. Beyond generating final predictions, LRMs can produce explicit reasoning chains. Many RL algorithms have been developed including, GRPO~\cite{guo2025deepseek}, REINFORCE++~\cite{hu2025reinforcestabilizingcriticfreepolicy}, DisCO~\cite{li2025disco}, to name just a few. 
Recent works~\cite{xu2025tabreason,cai2025tabr1} have applied RL directly to a base model for tabular prediction. However, reasoning capabilities of existing LRMs  do not readily transfer to tabular reasoning, which requires both semantic and numerical understanding~\cite{cai2025tabr1}.  As a result, directly generating reasoning traces from existing LRMs without fine-tuning on the tabular reasoning data may lead to hallucinations, undermining the faithfulness and trustworthiness of the outputs.
\cite{werling2025tables2traces} proposed an approach for curating reasoning traces from tabular data by constructing contrastive triplets of an example including the sample itself, the nearest samples with the same and different labels, and  prompting an generalist LLM to transform the triplets into contrastive decision processes, which are then used to supervise fine-tuning of a target LLM. However, this reasoning data is specifically constructed for medical question-answering tasks.  


{\bf Faithfulness of  Reasoning Models.} The reasoning traces generated from LRMs, although provide explicit explanation for the final prediction, could be unfaithful \cite{turpin_unfaithful, cot_in_the_wild_not_faithful, not_faithful_survey}. 
These works show that the reasoning chains of a model can be misleading or suffer from subtle biases that are not reflected in the traces, hampering the trustworthiness of the prediction. Thus, to ensure that the answer is trustworthy, evaluating the faithfulness of model reasoning explanation is imperative for any task, including tabular prediction. 
An important quality of a faithful reasoning is consistency, where important sentences or features in the reasoning trace have strong causal impact \cite{frodo, faithfulness_test} and replacing them with their counterfactual version should greatly affect the output of the model.
However, existing works that use LRMs in tabular prediction~\cite{xu2025tabreason, cai2025tabr1,werling2025tables2traces} only focus on evaluating prediction accuracy, lacking a comprehensive test to quantify the faithfulness quality of their models' reasoning.

\section{ReSS: Reasoning via Symbolic Scaffolds}
\label{sec:method}

{\bf Notations.} We consider classification problems on tabular data. Each sample can be represented by a tuple $(\x, y)$, where $\x$ is a feature vector $\mathbf{x} = (x_1, \ldots, x_d)^{\top} \in \R^d$ and $y \in \cY$ is the class label. The features in $\x$ may be numerical, ordinal, or categorical. Given an example $\x$, our goal is to fine-tune an LLM that given input features $\x$ predicts the label $y$, along with a reasoning trace $\mathbf{z} = (\z_1, \ldots, \z_k)$ to support and explain the final prediction, where each $\z_k$ represents a reasoning step consisting of a sequence of reasoning tokens. 

To input each tabular row into an LLM, we serialize the feature vector $\x$ into a sequence of natural-language tokens. Let us consider an example of diabetic prediction problem, where features may include age, BMI (Body Mass Index), DBP (Diastolic Blood Pressure), etc. Given a feature vector \{Age: 31, BMI:10.5, DBP: 66, \ldots \}, it is serialized to a  text, e.g.,  \textit{```The  age is 31,  the body mass index is 10.5, the diastolic blood pressure is 66, ..."}. 
Without causing any confusion, let $\x, \y$ denote the serialized texts for the input features and the output label, respectively.  Let $\pi_{\w}$ denote an LLM with parameters $\w$. We denote by $\pi_\w(\y|\x)$ the probability of generating $\y$ given $\x$. 

Before delving into the proposed approach, we first present two straightforward approaches for fine-tuning an LLM for tabular data prediction and discuss their limitations. 

\begin{figure*}[t]
  \begin{center}
\centerline{\includegraphics[width=0.85\textwidth]{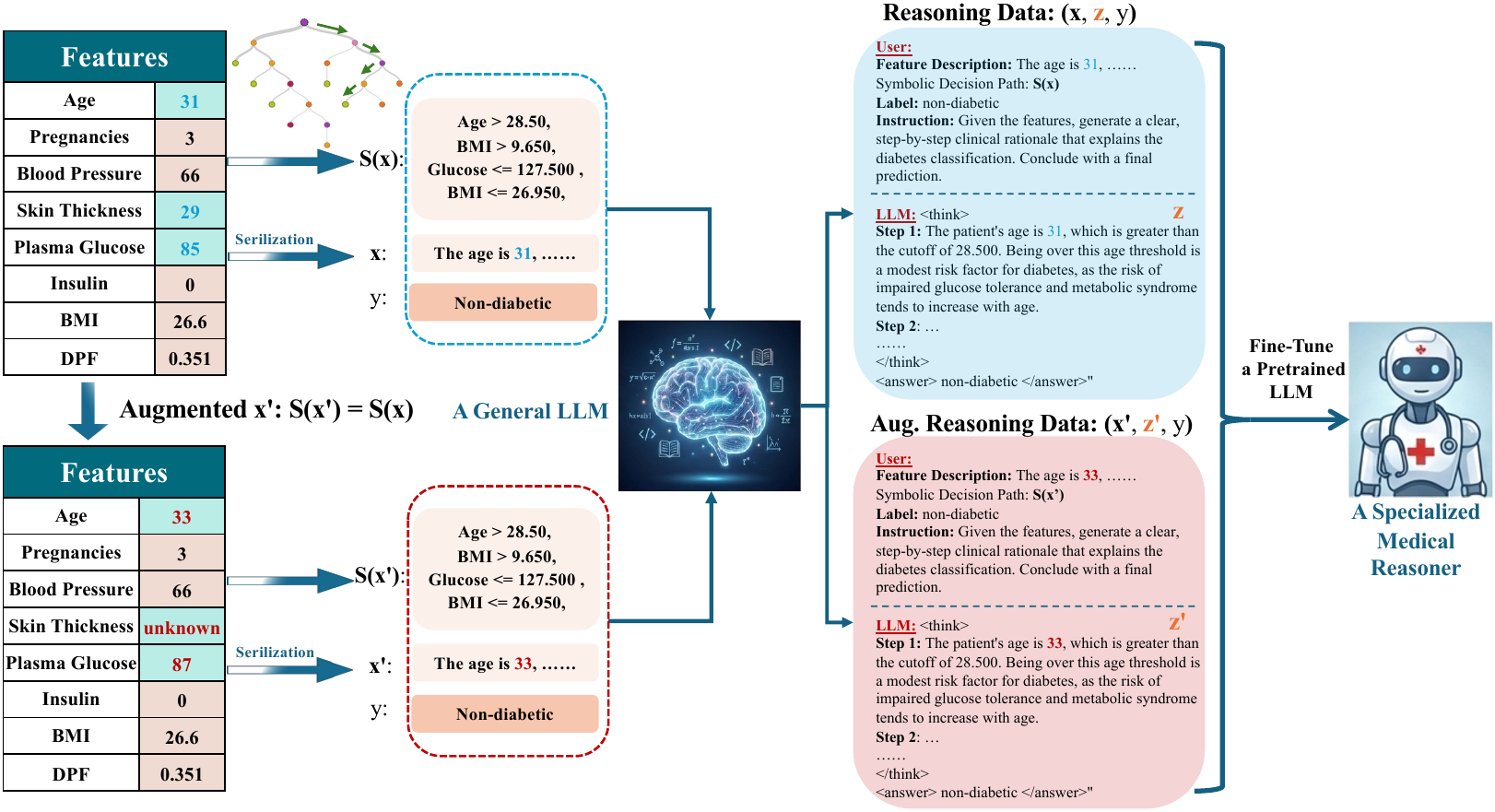}}
  \vspace*{-0.1in}
    \caption{
     An illustration of the ReSS pipeline applied to the diabetes prediction problem.
}
    \label{fig:overview}
  \end{center}
  \vspace*{-0.15in}
\end{figure*}
{\bf Baseline: Direct SFT.}  Given serialized data $\{(\x_i, \y_i)\}_{i=1}^n$, a straightforward approach to fine-tune an LLM is to perform the supervised fine-tuning (SFT) by minimizing the objective $ - \frac{1}{n}\sum_{i=1}^n\log \pi_{\w}(\y_i|\x_i, \t)$,
where $\t$ denote the task-related prompt. A limitation of this approach is that the model does not output any reasoning for the final prediction.

{\bf Baseline: Direct Reasoning Curation and SFT. }
One way to address the above limitation is to first curate a reasoning data and then conduct SFT.  However, curating high-quality reasoning data from domain experts in high-stakes domains is expensive, time-consuming, and hard to scale. A simple way is to leverage the knowledge of a general-purpose LLM. To this end, we can prompt an LLM with all features and the label $(\x, \y)$  and a guidance prompt $\t_c$ to generate a rationale for how a label $\y$ is predicted given the features in $\x$. We let $\mathbf R(\x, \y)$ denote the reasoning tokens generated by the LLM. 
Then, SFT is conducted by minimizing the objective $- \frac{1}{n}\sum_{i=1}^n\log \pi_{\w}(\{\z_i,\y_i\}|\x_i, \t)$,
where $\z_i =\mathbf R(\x_i, \y_i)$, $\{\z_i,\y_i\}$ denotes the concatenation of reasoning tokens and the final label tokens, and  $\t$ denotes the task-related prompt.

\subsection{Using Decision Tree Paths as Symbolic Scaffolds}

{\bf Motivation of Using Symbolic Scaffolds.}
A consequence of the direct curation approach discussed above is that the generated rationale may contain many non-useful features. An example given in Figure~\ref{fig:appendix:DRC_example_Diabetes} (Appendix~\ref{appendix:reasoning_example_DRC}) shows that the rationale generated by ChatGPT could iterate through all eight features on the diabetes dataset. To explain its limitation and motivate the proposed approach, we consider a simple example below. Let us consider the input data $\x\in\{1, -1\}^{2d}$, which contains a sequence of $\{1,-1\}$ tokens. Suppose the $y$ is given by $y = \prod_{i=1}^dx_{P(i)}$, where $P$ is a fixed (but unknown)  permutation of $\{1, \ldots, 2d\}$. Hence, $y$ is just the parity of $d$ fixed tokens in $\x$. The existing work~\cite{wies2022subtask} has shown that if we directly learn a model $\pi_{\w}(y|\x)$ to predict $y$ given $\x$, the learning requires an exponential number of samples in the order of $O(\exp(d))$. Hence, it is impossible to learn the correct model for predicting $y$ given $\x$ with limited data. The challenge of the problems lies that finding the correct permutation $P$ out of all permutations is exponentially hard. However, if we tell the learner that the final prediction $y$ only relies on the features in $x_{P(1)}, \ldots, x_{P(d)}$,  it could dramatically simplify the learning process.

This motivates the proposed approach of using {\bf symbolic scaffolds} to guide the feature selection and decision process. A symbolic scaffold refers to an explicit external structure such as rules, diagrams, decision trees, or step-by-step procedures that humans use to organize and guide their reasoning processes. In developmental psychology~\cite{Vygotsky1978Mind}, symbolic scaffolds are known to support problem solving by breaking complex tasks into interpretable intermediate steps, thereby reducing cognitive load and enabling systematic inference. Rather than replacing reasoning, these symbolic aids shape how reasoning unfolds by constraining the space of possible actions and making intermediate states explicit. {It is important to note that} the concept of a {\bf symbolic scaffold differs from that of chain-of-thought (CoT)}~\cite{wei2022chain} in reasoning models. CoT concerns how reasoning is expressed, whereas a symbolic scaffold determines how the reasoning process is structured and constrained.

To construct a symbolic scaffold for a tabular data, it is natural to use a decision tree model. To this end, we first train a decision tree on the given tabular data.  For each input instance $\mathbf{x}$, the tree induces a unique root-to-leaf symbolic decision path, which we formalize as an instance-specific symbolic
scaffold denoted by $S(\x)$. In particular, the scaffold $S(\x)$ is an ordered set of feature-level logical constraints that characterizes the sufficient conditions leading to the prediction for $\mathbf{x}$. Importantly, $S(\mathbf{x})$ is induced in a data-dependent manner: both the selected features and their associated thresholds arise from the empirical data distribution. As a result, the scaffold provides a compact, verifiable constraint space that can be directly used to guide rationale generation. An example of a scaffold $S(\x)$ from the decision tree is given in Figure~\ref{fig:overview}, ``age $>$26.5, BMI $>$9.625, ... ". Similar as before, we serialize $S(\x)$ into a sequence of tokens.

\subsection{Symbolic Scaffold Informed Reasoning Dataset Curation}
Given a symbolic scaffold that specifies the constraints of the decision process, we leverage the input features, the output label, and the corresponding symbolic scaffolds of correctly predicted instances to generate reasoning data. To this end, we  prompt a generalist LLM (e.g., ChatGPT) by asking it to provide the rationale  how the label $\y$ is predicted from the input features $\x$ following the  symbolic decision path $S(\x)$ that leads to the correct label.  Generalist LLMs encode rich domain knowledge acquired from large-scale pretraining, which can be leveraged to translate symbolic rules into coherent natural language explanations. In our setting, we use a generalist LLM as a constrained generator that injects domain knowledge while strictly adhering to symbolic constraints as specified in the symbolic scaffold. 

For each  example $(\x, \y)$ that is correctly predicted by the decision tree model, we let $\mathbf R(\x, \y, S(\x))$ denote the curated reasoning tokens. As a result, we collect a set of reasoning data $\{\x_i, \z_i, \y_i\}_{i\in\mathcal C}$, where $\z_i = \mathbf R(\x_i, \y_i, S(\x_i))$, and $\mathcal C\in[1,\ldots, n]$ denotes the subset of data that is correctly predicted by the decision tree model. 

\subsection{Scaffold-Invariant Data Augmentation}
\label{data_augmentation}
One might note that we only use the data that is correctly predicted by the decision tree model. This is important for LLM to learn consistent decision rules. However, it will shrink the size of training data for SFT. Below, we present an effective data augmentation strategy. 

A data augmentation is usually performed by perturbing the input features. However, a major challenge in data augmentation for tabular prediction is ensuring the consistency between perturbed features and the label, as unconstrained perturbations may cause an instance to no longer correspond to its original label. Our framework addresses this issue by enforcing {\bf scaffold invariance} during augmentation. 
Specifically, for each correctly predicted training instance by the decision tree model, its  symbolic scaffold $S(\x)$ characterizes the decision logic leading to its label. This scaffold defines a conjunction of feature-level constraints and serves as an explicit invariance during data augmentation. 

To generate an augmented sample, we randomly sample some features (e.g., by flipping a coin for each feature) and for each feature we randomly perturb its value in a valid set such that it does not violate the decision rule specified by the scaffold $S(\x)$. Perturbed values are sampled within a feature-specific valid range. To improve the out-of-distribution generalizability, we also allow for a certain probability (e.g., 0.2) to perturb a sampled feature that do not appear in the decision path  to the value \texttt{"unknown"}. This not only simulates out-of-distribution data but also facilitates the evaluation of explainability presented in next section. As a result,  all perturbations satisfy the original decision-path constraints, ensuring that the symbolic scaffold remains unchanged. By preserving the decision scaffold while allowing controlled feature-level variation, our augmentation procedure generates diverse yet label-scaffold consistent samples. 

For a perturbed sample $(\x', \y)$ corresponding to an original data point $(\x, \y)$, we follow the same procedure described in the previous subsection to generate the reasoning traces  $\z=\mathbf{R}(\x', \y, S(\x'))$. Let $\mathcal D=\{(\x_j, \z_j, \y_j)\}_{j=1}^m$ denote the final set of reasoning data including the original data and the perturbed data, we conduct SFT on the resulting reasoning data $\mathcal D$. 

\vspace*{0.1in}
\section{Evaluation of Faithfulness and Explainability}
\label{sec:Evaluation}
Since the ReSS-trained reasoning model is derived from an explainable and faithful decision tree model, a natural question is whether the learned reasoning model preserves the faithfulness and explainability of the underlying decision tree. To answer this question, we need to evaluate whether the model-generated reasoning is faithful and explainable. Without systematic evaluation, reasoning traces may appear plausible while relying on nonexistent features or implausible feature values, and including artificial rationales that are not causal for the final predictions.  To this end, we consider the following three complementary criteria: hallucination, explanation sufficiency, and explanation necessity, which together characterize whether the reasoning is grounded in the input, complete, and causally relevant to the prediction.

{\bf Hallucination.} We measure hallucination by identifying reasoning steps that violate the input evidence.
Specifically, we consider three types: (i) \emph{feature existence hallucination (FEH)}, where the reasoning refers to features absent from the input;
(ii) \emph{feature value hallucination (FVH)}, where incorrect or unsupported feature values are asserted;
and (iii) \emph{comparison hallucination (CH)}, where invalid numerical or categorical comparisons are made.
The hallucination rate is computed at the step level: for each reasoning step, we mark it as hallucinated if it contains a violation, and report the fraction of hallucinated steps among all steps.

{\bf Explanation Sufficiency.} To assess explanation sufficiency, we mask features that are not referenced in the generated reasoning traces by setting their values to \texttt{unknown} in the input, and prompting the model to re-generate reasoning and prediction, and evaluating the prediction accuracy.
Specifically, we progressively mask the number of such features from $k=1$ to $4$. If the accuracy remains stable under these perturbations, the original reasoning traces are considered to be sufficient. 

{\bf Explanation Necessity.} To assess explanation necessity, we mask the features explicitly appearing in the generated reasoning traces by setting their values to \texttt{unknown} in the input, and measure the resulting performance degradation.
Similarly, the number of masked features is varied from $k=1$ to $4$. A significant drop in accuracy suggests that the identified features are necessary for the prediction, while a minor change indicates redundant or spurious reasoning.

\section{Experiments}
\label{sec:experiments}
We conduct a comprehensive set of experiments to systematically evaluate the effectiveness of our proposed method.



{\bf Datasets.} We conduct experiments on five high-stakes tabular datasets from the medical and financial domains. In medical domain, we use an Alzheimer’s disease (AD) dataset with three classes: \textit{normal cognition, mild cognitive impairment, and Alzheimer’s disease}.~\cite{kwon2024largelanguagemodelsclinical}, and a Diabetes dataset to predict \textit{diabetes or non-diabetes}~\cite{hegselmann2023tabllmfewshotclassificationtabular}. In financial domain, we use the Creditg dataset to classify credit risk as \textit{good or bad}~\cite{hegselmann2023tabllmfewshotclassificationtabular}, and HomeLoan dataset to predict \textit{approved or not approved}  decisions sourced from Kaggle\footnote{https://www.kaggle.com/datasets/rishikeshkonapure/home-loan-approval}. Additionally, we incorporate a large-scale data \textit{Give Me Some Credit} (Credit-D) ~\cite{grinsztajn2022treebasedmodelsoutperformdeep} dataset to predict \textit{default or non-default} outcomes over a two-year period.

Detailed dataset statistics and data split information are summarized in Table~\ref{tab:combined}(left). For the AD dataset, we use its official split. For Credit-D, considering its larger scale, we perform a single split using a fixed random seed (Seed 0). For other three small datasets, we use random 80\%/10\%/10\% split for training, validation 
and testing, repeated three times.

To serialize tabular instances into natural-language text for LLMs, we adopt the Text Template proposed in~\cite{hegselmann2023tabllmfewshotclassificationtabular} for the Diabetes and Creditg datasets. 
For the AD and HomeLoan datasets, we design dataset-specific serialization templates, with details provided in Appendix~\ref{appendix:serialization}.
\begin{table*}[t]
\caption{
  \textbf{Left:} Dataset statistics and data split details.
  \textbf{Right:} Hallucination rates (\%) across different datasets. We report three types:
  Feature Existence Hallucination (FEH), Feature Value Hallucination (FVH), and Comparison
  Hallucination (CH). Results are averaged over three random seeds.
}
\label{tab:combined}
\centering
\begin{small}
\renewcommand{\arraystretch}{1.0}
\setlength{\tabcolsep}{4pt}
\begin{minipage}[t]{0.52\textwidth}
\centering
\begin{tabular}{lcccc|c}
\toprule
Dataset & \#Feature & Train & Val. & Test & Train$_\text{aug}$ \\
\midrule
HomeLoan  & 11 & 491   & 61   & 62   & 1624  \\
Diabetes  & 8  & 614   & 77   & 77   & 2144  \\
Creditg   & 20 & 800   & 100  & 100  & 2564  \\
AD        & 25 & 6062  & 303  & 759  & 8358  \\
Credit-D  & 10 & 13156 & 1644 & 1645 & 20340 \\
\bottomrule
\end{tabular}
\end{minipage}
\hspace{0.03\textwidth}
\begin{minipage}[t]{0.40\textwidth}
\centering
\begin{tabular}{lccc}
\toprule
Dataset & FEH (\%) & FVH (\%) & CH (\%) \\
\midrule
HomeLoan  & ${0}$ & $0$ & ${1.80}_{0.69}$ \\
Diabetes  & ${0}$ & $0$ & ${1.71}_{0.72}$ \\
Creditg   & ${0}$ & $0$ & ${0.85}_{0.22}$ \\
AD        & ${0}$ & $0$ & ${0.22}_{0.07}$ \\
Credit-D  & ${0}$ & $0$ & ${0.24}_{0.10}$ \\
\bottomrule
\end{tabular}
\end{minipage}
\end{small}
\vspace*{0.1in}
  \centering
  \caption{ Classification accuracy (\%) of different training methods across multiple datasets. Results are averaged over three random seeds for stochastic methods, with subscripts indicating the corresponding standard deviations. Decision Tree and XGBoost results on the AD and Credit-D datasets do not report standard deviations, as these methods are deterministic and only the fixed train–validation–test split is used.  }
\begin{tabular}{l|ccccc}
    \toprule
    Method & HomeLoan & Diabetes & Creditg & AD & Credit-D \\
    \midrule
    Decision Tree & $78.49_{0.93}$ & $76.62_{1.30}$ & $72.40_{3.72}$ & 73.12 & 77.45 \\
    XGboost & $77.42_{3.23}$ & $78.78_{3.75}$ & ${77.00}_{{3.00}}$ & 74.18 & 77.80 \\
    TabNet & $79.57_{1.52}$ & $77.49_{4.29}$ & $75.00_{0.33}$ & $74.70_{1.21}$ & $77.30_{0.63}$ \\
    TabPFN & $81.72_{0.76}$ & $76.19_{2.67}$ & $76.67_{3.09}$ & $74.92_{0.51}$ & $79.35_{0.08}$ \\
    \midrule
    Direct SFT ($\mathbf x\rightarrow \y$) & $80.32_{4.26}$ & $73.59_{2.21}$ & $71.20_{1.17}$ & $73.56_{1.36}$ & $77.63_{0.26}$ \\
    DRC + SFT ($\mathbf x\rightarrow \z\rightarrow \y$) & $77.96_{2.74}$ & $77.06_{6.48}$ & $74.00_{0.82}$ & $73.34_{1.02}$ & $77.08_{0.41}$ \\
    Direct RL ($\mathbf x\rightarrow \z\rightarrow \y$) & ${75.27}_{{4.62}}$ & ${82.25}_{{2.21}}$ & ${70.00}_{{0.00}}$ & ${74.75}_{{1.83}}$ & ${78.38}_{{0.39}}$ \\ 
    \midrule
    ReSS (w/o aug.; $\mathbf x\rightarrow\z \rightarrow \y$) & ${82.26}_{1.32}$ & ${81.39}_{{1.62}}$ & $76.67_{2.62}$ & ${75.80}_{{0.65}}$ & ${78.68}_{0.25}$ \\
    ReSS ($\mathbf x\rightarrow\z \rightarrow \y$) & $\mathbf{82.80}_\mathbf{2.01}$ & $\mathbf{85.71}_\mathbf{1.06}$ & $\mathbf{77.67}_\mathbf{1.88}$ & $\mathbf{76.07}_\mathbf{{0.06}}$ & $\mathbf{80.12}_\mathbf{0.10}$ \\
    \bottomrule
\end{tabular}

  \label{tab:accuracy}%
\vspace*{0.1in}

  \centering
  \caption{ Comparison of Different LLMs (each row corresponds to a single model). }
   \resizebox{\columnwidth}{!}{%
\begin{tabular}{l|ccccc}
    \toprule
    Method & Homeloan & Diabetes & Creditg & AD & Credit-D \\
    \midrule
    GPT4.1 ($\mathbf x\rightarrow \y$) & $81.18_{0.93}$ & $77.49_{0.75}$ & $35.00_{3.46}$ & 70.78 & 69.97 \\
    GPT4.1 ($\mathbf x\rightarrow \z \rightarrow \y$) & $80.64_{1.61}$ & $77.49_{1.98}$ & $57.33_{1.53}$ & $62.06$ & $71.82$ \\
    GPT4.1 ($\mathbf x, {S(\mathbf x)}\rightarrow \z \rightarrow \y$) & $80.11_{0.93}$ & $74.89_{0.75}$ & $60.33_{1.53}$ & $70.22$ & $72.04$ \\
    Qwen-2.5-3B-Instruct ($\mathbf x\rightarrow \y$) & $51.08_{2.74}$ & $70.13_{1.06}$ & $30.00_{0.00}$ & 64.94 & 34.53 \\
    ReSS-Qwen-2.5-3B-Instruct ($\mathbf x\rightarrow\z \rightarrow \y$) & ${\bf 81.18}_{0.76}$ & ${\bf 81.82}_{2.81}$ & ${\bf 76.67}_{2.62}$ & ${\bf 74.35}_{0.35}$ & ${\bf 79.98}_{0.20}$ \\
    \bottomrule
\end{tabular}
    }
    
  \label{tab:accuracy-llm}%
\end{table*}%


{\bf Baselines.} We compare our method against a diverse set of training baselines covering classical machine learning, deep learning, and LLM-based approaches.
Specifically, we include traditional machine learning methods (1) Decision Tree~\cite{DBLP:books/wa/BreimanFOS84} and (2) XGBoost\cite{Chen_2016}; (3) TabNet~\cite{arik2020tabnetattentiveinterpretabletabular}, a representative deep learning method for tabular prediction; (4) TabPFN~\cite{hollmann2023tabpfntransformersolvessmall}, a trained Transformer to approximate probabilistic inference for tabular classification tasks. For baselines involving fine-tuning LLMs, we further include (5) Direct SFT and (6) Direct Reasoning Curation  followed by SFT (DRC+SFT); (7) Direct RL approach that directly conducts RL on the base LLM to exploit the existing reasoning capabilities of the base LLM, corresponding to the approach in~\cite{xu2025tabreason}. For all LLM fine-tuning methods, we use Qwen-2.5-3B-Instruct\footnote{Released under Apache License 2.0.} as the base model. For RL, we use the recently proposed DisCO algorithm~\cite{li2025disco}, which has been observed to be better than GRPO.

{\bf Implementation Details.} We first train a decision tree  with a grid search  on each dataset and then follow the procedure of ReSS to curate the reasoning data and fine-tune an LLM. We use GPT-4.1 to generate the reasoning data. Due to space limit, we include the exact prompts used for different methods during data curation, SFT and RL are provided in Appendix~\ref{appendix:prompt}. Curated reasoning examples by ReSS and the baseline DRC are shown in Appendix~\ref{appendix:reasoning_example}. All hyperparameter settings and compute resources are detailed in Appendix~\ref{appendix:hyperparameters}. For ReSS on Creditg, Diabetes, and HomeLoan, we generate augmented samples equal to three times the number of correctly predicted instances using a decision tree, while for the larger AD and Credit-D datasets, we generate one additional copy.  We report the final effective training size including augmented samples in the column Train$_\text{aug}$ in Table~\ref{tab:combined} (left).  Note that except for AD and Credit-D, these numbers are averaged over three random seeds.


\subsection{Comparison with Baselines}
\label{sec:baseline_comparison}

Table~\ref{tab:accuracy} summarizes the classification accuracy of different methods across five tabular datasets. 
We can observe that Direct SFT generally underperforms traditional methods. Introducing additional reasoning supervision (DRC+SFT) improves performance on some datasets but yields inconsistent gains, indicating the limitations of direct unconstrained reasoning curation. In contrast, with symbolic scaffold informed reasoning boosted by the scaffold-invariant data augmentation, ReSS consistently outperforms all baselines, with particularly notable gains on Diabetes and AD datasets. 
We also see that Direct RL approach shows high variance across datasets, reflecting the difficulty of optimizing reasoning behavior without explicit structural guidance. These results collectively demonstrate that benefits of structured reasoning supervision grounded in symbolic scaffolds, underscoring the importance of high-quality, structure-aligned reasoning data for effective tabular decision-making.

\subsection{Evaluation of Faithfulness and Explainability}

As discussed in Section~\ref{sec:Evaluation}, we evaluate faithfulness and explainability on our ReSS models trained with the augmented datasets from three complementary perspectives:
\emph{hallucination}, \emph{explanation sufficiency}, and \emph{explanation necessity}.

{\bf Hallucination Analysis.}
Table~\ref{tab:combined}(right) reports the hallucination rates across five datasets. 
Across all datasets, FEH and FVH are consistently zero, indicating that the model does not invent
non-existent features or incorrect feature values in its reasoning.
Comparison Hallucination occurs only rarely, with rates below $2\%$ on all datasets.
These results demonstrate that the generated explanations are highly faithful to the underlying input features,
with only occasional errors arising from incorrect relational comparisons.

\begin{figure*}[h]
  \begin{center}
\centerline{\includegraphics[width=\textwidth]{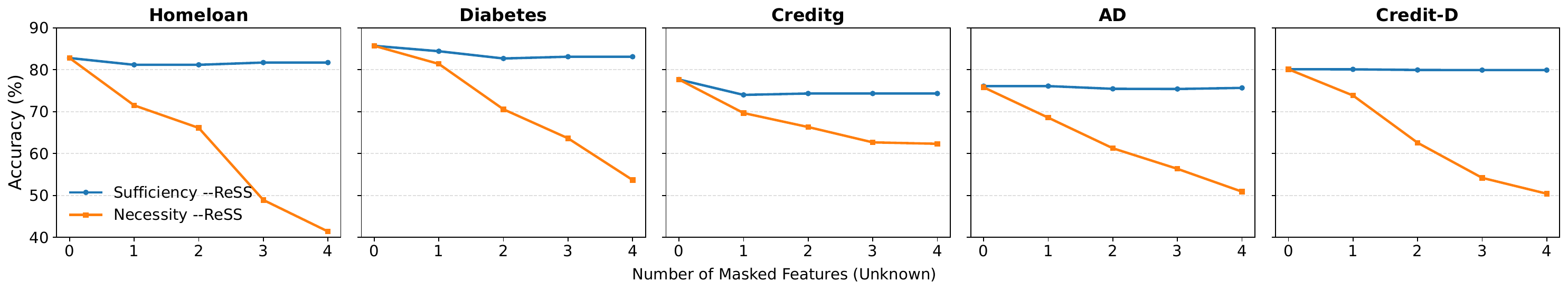}}
      \vspace*{-0.1in}
\caption{
     Explanation sufficiency and necessity analysis for ReSS via feature masking across five tabular datasets, averaged over three random seeds. The x-axis denotes the number of masked features per instance, while the y-axis shows the resulting change in prediction accuracy under masking interventions.}
    \label{fig:sufficiency_necessity}
  \end{center}
  \vspace*{-0.25in}
\end{figure*}

{\bf Explanation Sufficiency and Necessity.}
Figure~\ref{fig:sufficiency_necessity} illustrates the results of explanation sufficiency and necessity evaluation
by progressively masking features with an \texttt{unknown} token.
We observe that across all datasets, masking unused features in reasoning has only a marginal impact on performance,
indicating that the explanations are sufficient: they already capture the core information needed for prediction.
In contrast, masking used features in reasoning leads to a substantial and monotonic drop in accuracy, where performance degrades sharply as more explanatory features are removed.
This clear divergence between the two curves demonstrates that the explanations rely on features that are genuinely
necessary for the model’s decision, providing strong evidence of both explanation sufficiency and necessity.
We further analyze explainability under the absence of symbolic guidance by comparing ReSS with DRC+SFT, and observe a substantial degradation in explanation necessity without symbolic scaffolds; detailed results are provided in Appendix~\ref{appendix:reasoning_explainability}.

\subsection{Comparison with Different LLMs}
While LLMs are generally regarded as general-purpose models, the models reported in Table~\ref{tab:accuracy} are trained separately for each dataset. This raises a natural question: can a single model be trained to perform prediction across all datasets? To investigate this, we train a unified model, termed ReSS-Qwen-2.5-3B-Instruct, using the union of all augmented reasoning data from all datasets, and evaluate its performance on each individual dataset.

Table~\ref{tab:accuracy-llm} compares our fine-tuned unified model with different LLMs, including GPT-4.1 and  Qwen-2.5-3B-Instruct (our base model). For GPT-4.1, we evaluate two settings: (i) a direct prediction setting, where the model directly predicts the label given the input features and prompt, denoted by $(\mathbf{x}\!\rightarrow\!\mathbf{y})$; (ii) a reasoning-based setting, where the model is first prompted to generate explicit reasoning (using prompts consistent with our reasoning curation prompts) and then predict the label, denoted by $(\mathbf{x}\!\rightarrow\!\mathbf{z}\!\rightarrow\!\mathbf{y})$; and (iii) a scaffold-augmented reasoning setting, where the symbolic decision path $S(\mathbf{x})$ extracted from the trained decision tree is additionally provided as part of the input prompt, denoted by $(\mathbf{x}, S(\mathbf{x})\!\rightarrow\!\mathbf{z}\!\rightarrow\!\mathbf{y})$. The prompts are provided in ~\ref{appendix:prompt_GPT_reasoning}, ~\ref{appendix:prompt_GPT_wo_reasoning} and ~\ref{appendix:prompt_GPT_w_scaffold_reasoning}.

Despite its strong general capabilities, GPT-4.1 shows unreliable performance on tabular prediction tasks under both $(\mathbf{x}\!\rightarrow\!\mathbf{y})$ and $(\mathbf{x}\!\rightarrow\!\mathbf{z}\!\rightarrow\!\mathbf{y})$, suggesting that general ability does not transfer well to structured tabular decision-making. Notably, supplying the decision path $S(\mathbf{x})$ at inference time does not yield consistent improvement. This shows that symbolic scaffolds are not effective as raw inference-time prompts and must be internalized through fine-tuning. In contrast, ReSS-Qwen-2.5-3B-Instruct consistently achieves higher accuracy across datasets, indicating that effective tabular reasoning relies on dataset-aligned reasoning supervision rather than on model scale or scaffold availability alone.


\begin{figure}[h]
  \centering
    \hspace*{-0.15in}
  \includegraphics[width=0.41\columnwidth]{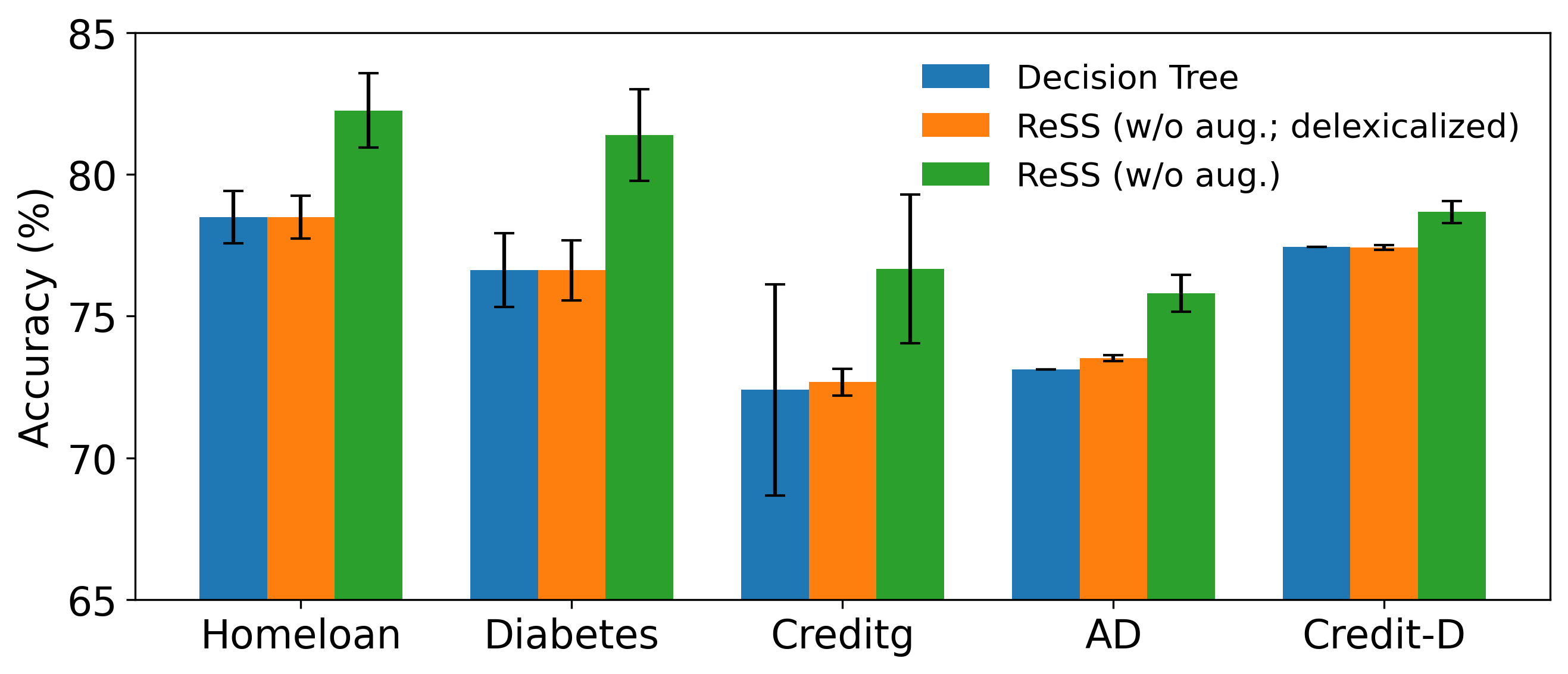}
  \hspace*{-0.05in}
  \includegraphics[width=0.6\columnwidth]{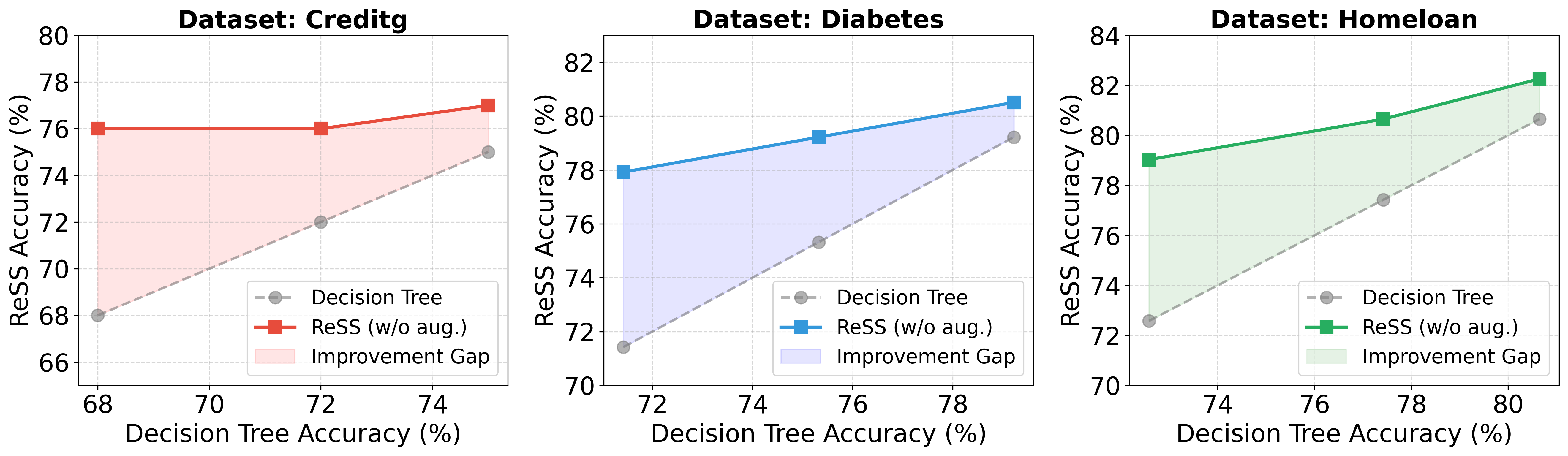}
  \vspace*{-0.1in}
  \caption{
    \textbf{Left:} Ablation study with delexicalized features, conducted without augmented reasoning data.
    \textbf{Right:} Sensitivity to decision tree qualities. 
    Results are averaged over three random seeds.
  }
  \label{fig:combined}
\end{figure}

\subsection{Ablation studies}
{\bf Domain Knowledge Matters.}
To quantify the role of domain knowledge, we conduct an ablation study that delexicalizes all feature names and values during reasoning data curation. Specifically, we preserve the original input structure—including the input features, symbolic scaffolds, and the label—but anonymize all feature names and categorical values into abstract placeholders (e.g., \textit{Feature A}, \textit{Category B}). The LLM is strictly instructed to follow the symbolic scaffold while being prohibited from providing any semantic or domain-specific explanations, resulting in reasoning that purely imitates the symbolic structure without introducing external domain knowledge. Prompts for reasoning data curation and SFT, and curated reasoning examples, are provided in Appendix~\ref{appendix:prompt_delexicalized} and~\ref{appendix:reasoning_example_delexicalized_ReSS}, respectively.

Figure~\ref{fig:combined} (left) shows that models trained on delexicalized reasoning achieve performance nearly identical to the decision tree baseline across all datasets, but consistently underperform ReSS. This result indicates that delexicalized reasoning mainly captures the symbolic structure of the decision tree, while the additional performance gains of ReSS arise from domain knowledge introduced by a general LLM. We include a cherry-picked analysis in Appendix~\ref{appendix:cherry_pick} to illustrate how domain knowledge contributes to the performance gains of our ReSS-trained model over decision trees .

{\bf Sensitivity to Decision Tree Quality.}
A natural concern is whether ReSS critically relies on a 
high-quality base decision tree. To examine this, we vary the 
accuracy of the decision tree learend with different hyper-parameters and report the corresponding 
ReSS performance on \textit{Creditg},  \textit{Diabetes} and \textit{Homeloan}, 
as shown in Figure~\ref{fig:combined} (right).
Across both datasets, ReSS consistently outperforms the baseline 
by a clear margin. Even with a weak decision tree (e.g., $68\%$ 
on \textit{Creditg} and $71.42\%$ on \textit{Diabetes}), ReSS 
still attains $76\%$ and $77.92\%$, respectively. Moreover, the 
final ReSS accuracy varies only marginally across different 
decision tree qualities, while the \textit{improvement gap} 
(shaded region) is largest when the baseline is weakest. 
This indicates that ReSS is robust to the quality of the 
input decision tree: it does not require a near-optimal tree 
and can extract useful structural signals even from relatively 
weak ones, making it practical in real-world settings.

{\bf SFT vs RL.} 
Since RL is usually regarded as an effective approach to improve the generalization of SFT, we conduct RL on ReSS-trained models. To this end, we conduct RL based on DisCO from the model initialized from ReSS-trained model without augmentation.  We compare this approach, denoted by ReSS (w/o aug) + RL, with the direct RL approach in Figure~\ref{fig:ReSS_RL_Direct_RL} (Appendix~\ref{app:SFTRL}), which shows that performing RL after ReSS  consistently outperforms directly applying RL to the base model across all datasets, demonstrating that RL benefits substantially from structured reasoning supervision. We also compare the approach ReSS (w/o aug) + RL with ReSS  and ReSS (w/o aug.) in Figure~\ref{fig:RL_aug_Ress} (Appendix~\ref{app:SFTRL}). We observe that while RL generally improves its SFT counterparts, it is not necessarily better than ReSS-trained model on augmented data.  This indicates that our scaffold-invariant data augmentation is an effective approach for improving generalization.
We additionally apply RL on top of ReSS with augmented data. However, this setting does not yield further consistent improvements. This suggests that scaffold-invariant augmentation already supplies a strong learning signal, diminishing the marginal benefits of additional RL.

\section{Conclusions}
We proposed ReSS, a framework for learning faithful reasoning models for tabular data by using decision-tree paths as symbolic scaffolds. ReSS enables scalable reasoning data curation and trains LLMs to produce structured, causally grounded explanations. Experiments on medical and financial datasets demonstrate improved predictive performance and strong faithfulness and explainability. A limitation of this study is its focus on financial and medical domains. Consequently, the effectiveness of ReSS in other areas where LLMs possess significant intrinsic knowledge remains unverified.



\bibliography{example_paper}

\begin{thebibliography}{10}

\bibitem{abutbul2020dnfnetneuralarchitecturetabular}
Ami Abutbul, Gal Elidan, Liran Katzir, and Ran El-Yaniv.
\newblock Dnf-net: A neural architecture for tabular data, 2020.

\bibitem{cot_in_the_wild_not_faithful}
Iv{\'a}n Arcuschin, Jett Janiak, Robert Krzyzanowski, Senthooran Rajamanoharan, Neel Nanda, and Arthur Conmy.
\newblock Chain-of-thought reasoning in the wild is not always faithful, 2025.
\newblock {\em URL https://arxiv. org/abs/2503.08679}, 2025.

\bibitem{arik2020tabnetattentiveinterpretabletabular}
Sercan~O. Arik and Tomas Pfister.
\newblock Tabnet: Attentive interpretable tabular learning, 2020.

\bibitem{faithfulness_test}
Pepa Atanasova, Oana-Maria Camburu, Christina Lioma, Thomas Lukasiewicz, Jakob~Grue Simonsen, and Isabelle Augenstein.
\newblock Faithfulness tests for natural language explanations.
\newblock {\em arXiv preprint arXiv:2305.18029}, 2023.

\bibitem{not_faithful_survey}
Fazl Barez, Tung-Yu Wu, Iv{\'a}n Arcuschin, Michael Lan, Vincent Wang, Noah Siegel, Nicolas Collignon, Clement Neo, Isabelle Lee, Alasdair Paren, et~al.
\newblock Chain-of-thought is not explainability.
\newblock {\em Preprint, alphaXiv}, page~v1, 2025.

\bibitem{10.1023/A:1010933404324}
Leo Breiman.
\newblock Random forests.
\newblock {\em Mach. Learn.}, 45(1):5–32, October 2001.

\bibitem{DBLP:books/wa/BreimanFOS84}
Leo Breiman, J.~H. Friedman, Richard~A. Olshen, and C.~J. Stone.
\newblock {\em Classification and Regression Trees}.
\newblock Wadsworth, 1984.

\bibitem{cai2025tabr1}
Pengxiang Cai, Zihao Gao, and Jintai Chen.
\newblock Tabr1: Taming grpo for tabular reasoning llms.
\newblock {\em arXiv preprint arXiv:2510.17385}, 2025.

\bibitem{Chen_2016}
Chen, Tianqi, Guestrin, and Carlos.
\newblock Xgboost: A scalable tree boosting system.
\newblock In {\em Proceedings of the 22nd ACM SIGKDD International Conference on Knowledge Discovery and Data Mining}, KDD ’16, page 785–794. ACM, August 2016.

\bibitem{diabetes_130-us_hospitals_for_years_1999-2008_296}
John Clore, Krzysztof Cios, Jon DeShazo, and Beata Strack.
\newblock {Diabetes 130-US Hospitals for Years 1999-2008}.
\newblock UCI Machine Learning Repository, 2014.
\newblock {DOI}: 10.24432/C5230J.

\bibitem{dinh2022liftlanguageinterfacedfinetuningnonlanguage}
Tuan Dinh, Yuchen Zeng, Ruisu Zhang, Ziqian Lin, Michael Gira, Shashank Rajput, Jy~yong Sohn, Dimitris Papailiopoulos, and Kangwook Lee.
\newblock Lift: Language-interfaced fine-tuning for non-language machine learning tasks, 2022.

\bibitem{grinsztajn2022treebasedmodelsoutperformdeep}
Léo Grinsztajn, Edouard Oyallon, and Gaël Varoquaux.
\newblock Why do tree-based models still outperform deep learning on tabular data?, 2022.

\bibitem{guo2025deepseek}
Daya Guo, Dejian Yang, Haowei Zhang, Junxiao Song, Ruoyu Zhang, Runxin Xu, Qihao Zhu, Shirong Ma, Peiyi Wang, Xiao Bi, et~al.
\newblock Deepseek-r1: Incentivizing reasoning capability in llms via reinforcement learning.
\newblock {\em arXiv preprint arXiv:2501.12948}, 2025.

\bibitem{hegselmann2023tabllmfewshotclassificationtabular}
Stefan Hegselmann, Alejandro Buendia, Hunter Lang, Monica Agrawal, Xiaoyi Jiang, and David Sontag.
\newblock Tabllm: Few-shot classification of tabular data with large language models, 2023.

\bibitem{hollmann2023tabpfntransformersolvessmall}
Noah Hollmann, Samuel Müller, Katharina Eggensperger, and Frank Hutter.
\newblock Tabpfn: A transformer that solves small tabular classification problems in a second, 2023.

\bibitem{hu2025reinforcestabilizingcriticfreepolicy}
Jian Hu, Jason~Klein Liu, Haotian Xu, and Wei Shen.
\newblock Reinforce++: Stabilizing critic-free policy optimization with global advantage normalization, 2025.

\bibitem{huang2020tabtransformertabulardatamodeling}
Xin Huang, Ashish Khetan, Milan Cvitkovic, and Zohar Karnin.
\newblock Tabtransformer: Tabular data modeling using contextual embeddings, 2020.

\bibitem{https://doi.org/10.1111/nyas.15339}
Subbarao Kambhampati, Kaya Stechly, and Karthik Valmeekam.
\newblock (how) do reasoning models reason?
\newblock {\em Annals of the New York Academy of Sciences}, 1547(1):33--40, 2025.

\bibitem{NIPS2017_6449f44a}
Guolin Ke, Qi~Meng, Thomas Finley, Taifeng Wang, Wei Chen, Weidong Ma, Qiwei Ye, and Tie-Yan Liu.
\newblock Lightgbm: A highly efficient gradient boosting decision tree.
\newblock In I.~Guyon, U.~Von Luxburg, S.~Bengio, H.~Wallach, R.~Fergus, S.~Vishwanathan, and R.~Garnett, editors, {\em Advances in Neural Information Processing Systems}, volume~30. Curran Associates, Inc., 2017.

\bibitem{kwon2024largelanguagemodelsclinical}
Taeyoon Kwon, Kai~Tzu iunn Ong, Dongjin Kang, Seungjun Moon, Jeong~Ryong Lee, Dosik Hwang, Yongsik Sim, Beomseok Sohn, Dongha Lee, and Jinyoung Yeo.
\newblock Large language models are clinical reasoners: Reasoning-aware diagnosis framework with prompt-generated rationales, 2024.

\bibitem{li2025disco}
Gang Li, Ming Lin, Tomer Galanti, Zhengzhong Tu, and Tianbao Yang.
\newblock Disco: Reinforcing large reasoning models with discriminative constrained optimization.
\newblock {\em arXiv preprint arXiv:2505.12366}, 2025.

\bibitem{bank_marketing_222}
S.~Moro, P.~Rita, and P.~Cortez.
\newblock {Bank Marketing}.
\newblock UCI Machine Learning Repository, 2014.
\newblock {DOI}: 10.24432/C5K306.

\bibitem{frodo}
Debjit Paul, Robert West, Antoine Bosselut, and Boi Faltings.
\newblock Making reasoning matter: Measuring and improving faithfulness of chain-of-thought reasoning.
\newblock {\em arXiv preprint arXiv:2402.13950}, 2024.

\bibitem{si2024interpretabnet}
Jacob Si, Wendy~Yusi Cheng, Michael Cooper, and Rahul~G Krishnan.
\newblock Interpretabnet: Distilling predictive signals from tabular data by salient feature interpretation.
\newblock {\em arXiv preprint arXiv:2406.00426}, 2024.

\bibitem{slack2023tabletlearninginstructionstabular}
Dylan Slack and Sameer Singh.
\newblock Tablet: Learning from instructions for tabular data, 2023.

\bibitem{team2025kimi}
Kimi Team, Angang Du, Bofei Gao, Bowei Xing, Changjiu Jiang, Cheng Chen, Cheng Li, Chenjun Xiao, Chenzhuang Du, Chonghua Liao, et~al.
\newblock Kimi k1. 5: Scaling reinforcement learning with llms.
\newblock {\em arXiv preprint arXiv:2501.12599}, 2025.

\bibitem{turpin_unfaithful}
Miles Turpin, Julian Michael, Ethan Perez, and Samuel Bowman.
\newblock Language models don't always say what they think: Unfaithful explanations in chain-of-thought prompting.
\newblock {\em Advances in Neural Information Processing Systems}, 36:74952--74965, 2023.

\bibitem{vonwerra2022trl}
Leandro von Werra, Younes Belkada, Lewis Tunstall, Edward Beeching, Tristan Thrush, Nathan Lambert, Shengyi Huang, Kashif Rasul, and Quentin Gallouédec.
\newblock Trl: Transformer reinforcement learning.
\newblock \url{https://github.com/huggingface/trl}, 2020.

\bibitem{Vygotsky1978Mind}
L.~S. Vygotsky.
\newblock {\em {Mind in Society: Development of Higher Psychological Processes}}.
\newblock Harvard University Press, 14th edition, March 1978.

\bibitem{wei2022chain}
Jason Wei, Xuezhi Wang, Dale Schuurmans, Maarten Bosma, brian ichter, Fei Xia, Ed~H. Chi, Quoc~V Le, and Denny Zhou.
\newblock Chain of thought prompting elicits reasoning in large language models.
\newblock In Alice~H. Oh, Alekh Agarwal, Danielle Belgrave, and Kyunghyun Cho, editors, {\em Advances in Neural Information Processing Systems}, 2022.

\bibitem{werling2025tables2traces}
Mikkel Werling, Nabeel Seedat, Jiashuo Liu, Lars Gr{\o}nlykke, Carsten~Utoft Niemann, Mihaela van~der Schaar, and Rudi Agius.
\newblock Tables2traces: Distilling tabular data to improve llm reasoning in healthcare.
\newblock In {\em EurIPS 2025 Workshop: AI for Tabular Data}, 2025.

\bibitem{wies2022subtask}
Noam Wies, Yoav Levine, and Amnon Shashua.
\newblock Sub-task decomposition enables learning in sequence to sequence tasks.
\newblock In {\em International Conference on Learning Representations}, 2023.

\bibitem{xu2025tabreason}
Tommy Xu, Zhitian Zhang, Xiangyu Sun, Lauren~Kelly Zung, Hossein Hajimirsadeghi, and Greg Mori.
\newblock Tabreason: A reinforcement learning-enhanced reasoning llm for explainable tabular data prediction.
\newblock {\em arXiv preprint arXiv:2505.21807}, 2025.

\end{thebibliography}
\bibliographystyle{plain}
\newpage
\appendix
\onecolumn

\section{Tabular Serialization}
\label{appendix:serialization}
\subsection{Alzheimer's Disease Dataset}
Each patient record is serialized into a single natural-language paragraph by applying a deterministic, rule-based feature-to-text mapping.  
Let $\mathbf{x}$ denote a structured feature vector for one subject.  
We generate a textual note by iterating over all features in $\mathbf{x}$ and applying the following serialization rules.

\begin{itemize}
    \item \textbf{Atrophy-related brain regions.}  
    For features corresponding to regional brain atrophy (e.g., hippocampal, entorhinal, parietal lobe), each feature $f$ with value $v$ is serialized as:
    \begin{quote}
        \emph{``The atrophy level of $f$ is $v$.''}
    \end{quote}

    \item \textbf{Non-atrophy neuroimaging features.}  
    Neuroimaging indicators that do not represent regional atrophy are serialized using feature-specific clinically appropriate phrasing, for example:
    \begin{itemize}
        \item \emph{Overall (cortex)} $\rightarrow$ ``The overall cortical atrophy is $v$.''  
        \item \emph{Ventricle enlargement} $\rightarrow$ ``The level of ventricle enlargement is $v$.''  
        \item \emph{Lateral ventricle shape} $\rightarrow$ ``The temporal/frontal shape of the lateral ventricle is $v$.''  
        \item \emph{Vascular disease} $\rightarrow$ ``The evidence of vascular disease is $v$.''  
    \end{itemize}

    \item \textbf{Demographic and clinical variables.}  
    Demographic and non-imaging clinical features are serialized as direct factual statements, for example:
    \begin{itemize}
        \item \emph{Age} $\rightarrow$ ``The age is $v$ years old.''  
        \item \emph{Sex} $\rightarrow$ ``The sex is $v$.''  
        \item \emph{MMSE score} $\rightarrow$ ``The MMSE score is $v$.''  
        \item \emph{APOE $\varepsilon4$ status} $\rightarrow$ ``The APOE $\varepsilon4$ status is $v$.''  
        \item \emph{Education} $\rightarrow$ ``The years of education are $v$.''  
        \item \emph{Marital status} $\rightarrow$ ``The marital status is $v$.''  
    \end{itemize}

\end{itemize}

\subsection{Homeloan Dataset}

For the Home Loan dataset, each applicant record is serialized into a single natural-language paragraph by converting each tabular feature into a fixed factual sentence template.  
The serialization is deterministic and applied independently to each feature.

The following feature-specific templates are used:

\begin{itemize}
    \item \textbf{Gender}:  
    ``The gender of the applicant is $v$.''

    \item \textbf{Married}:  
    ``The marital status of the applicant is $v$.''

    \item \textbf{Dependents}:  
    ``The number of dependents is $v$.''

    \item \textbf{Education}:  
    ``The education level of the applicant is $v$.''

    \item \textbf{Self\_Employed}:  
    ``The self-employment status is $v$.''

    \item \textbf{ApplicantIncome}:  
    ``The applicant income is $v$.''

    \item \textbf{CoapplicantIncome}:  
    ``The co-applicant income is $v$.''

    \item \textbf{LoanAmount}:  
    ``The loan amount is $v$.''

    \item \textbf{Loan\_Amount\_Term}:  
    ``The loan amount term is $v$.''

    \item \textbf{Credit\_History}:  
    ``The credit history value is $v$.''

    \item \textbf{Property\_Area}:  
    ``The property area is $v$.''
\end{itemize}

\subsection{Credit-D Dataset}

For the Credit-D dataset, each applicant record is serialized into a single natural-language paragraph by converting each tabular feature into a fixed factual sentence template.  
The serialization is deterministic and applied independently to each feature.

The following feature-specific templates are used:

\begin{itemize}
\item \textbf{RevolvingUtilizationOfUnsecuredLines}:
``The revolving utilization of unsecured lines is $v$.''
\item \textbf{age}: 
``The age of the applicant is $v$.''

\item \textbf{NumberOfTime30-59DaysPastDueNotWorse}: 
``The number of times the applicant was 30 to 59 days past due without worse delinquency is $v$.''

\item \textbf{DebtRatio}: 
``The debt ratio is $v$.''

\item \textbf{MonthlyIncome}: 
``The monthly income is $v$.''

\item \textbf{NumberOfOpenCreditLinesAndLoans}: 
``The number of open credit lines and loans is $v$.''

\item \textbf{NumberOfTimes90DaysLate}: 
``The number of times the applicant was 90 days late is $v$.''

\item \textbf{NumberRealEstateLoansOrLines}: 
``The number of real estate loans or lines is $v$.''

\item \textbf{NumberOfTime60-89DaysPastDueNotWorse}: 
``The number of times the applicant was 60 to 89 days past due without worse delinquency is $v$.''

\item \textbf{NumberOfDependents}: 
``The number of dependents is $v$.''
\end{itemize}

\section{Prompts}
\label{appendix:prompt}
\subsection{Prompts for ReSS}
\subsubsection{Reasoning Data Curation}
\label{appendix:prompt_Ress}
\textbf{Alzheimer's Disease Dataset}
\begin{small}
\begin{verbatim}
You are a medical reasoning assistant for Alzheimer's monitoring.
Generate a step-by-step rationale that follows exactly the conditions on the given 
decision rule path.
Do NOT mention terms like "decision tree", "splitting", or "branches".

Instructions:
- Only include the features that appear on the decision path.
- Keep the order consistent with the decision path sequence.
- Do not add extra features that are not used on the path.

For each step:
- First, state a purely factual clinical description of the patient based on the 
  given condition:
  * For continuous numeric clinical features, explicitly state the exact measured value 
    and whether it is <= or > the specified cutoff.
  * For categorical clinical features, state only the real-world clinical category, 
    without mentioning any numeric values, indicators, encodings, or thresholds.
  * For ordinal clinical features, explicitly state the true ordinal value first 
    (0, 1, or 2) as observed in the data, and then immediately provide its clinical 
    interpretation according to the interpret-feature definition.
- Then, explain the clinical implication of this finding in clear Alzheimer's 
  disease-related diagnostic terms.

In the final synthesis:
- Do not state or imply the final classification in the first sentence.
- First summarize the evidence balance:
  * List which steps support each class (Normal Cognition / Mild Cognitive Impairment / 
    Alzheimer's Disease).
  * Some steps may be neutral and support no class.
  * Explain how the distribution and strength of supporting evidence across steps leads 
    to one class outweighing the others.
  * The final conclusion must be based solely on the clinical meaning of the evidence 
    described in each step, and must not rely on the decision rule structure.
- Only AFTER that, state the final classification as exactly one of the allowed classes.

Interpret features:
- Brain regions: 0=normal, 1=mild atrophy, 2=severe atrophy.
- Lateral Ventricle shape: 0=round, 1=moderate change, 2=sharp (abnormal).
- Vascular disease: 0=no, 1=yes.
- MMSE: lower = worse cognition.
- APOE4: 0=no allele, 1=one allele, 2=two alleles (higher risk).
- Education: higher = more cognitive reserve.
- Age: higher age increases Alzheimer's risk; younger age makes significant 
  neurodegeneration less likely.

Input:
Patient data: {patient_features}
Decision rule path: {decision_path}
Classification: {label}

Output format:
<think>
Step 1: ...
Step 2: ...
...
Overall, [final synthesis], these characteristics collectively support a "{label}" 
classification.
</think>
<answer> {label} </answer>
\end{verbatim}
\end{small}

\textbf{Creditg Dataset}
\begin{small}
\begin{verbatim}
You are a credit risk assessment assistant.
Your task is to generate a clear, step-by-step rationale that follows exactly the 
conditions on the given decision rule path.
Do NOT mention terms like "decision tree", "splitting", or "branches".

Instructions:
- Use only the features that appear in the decision path.
- For each step, describe it purely as a factual statement about the applicant -- for 
  continuous numeric features, explicitly state the exact value and whether it is <= or >
  the cutoff; for categorical features, state only the real-world category without
  mentioning any numeric values, indicators, encodings, or thresholds -- and then explain
  what this condition implies for the applicant's credit risk.
- Keep the steps ordered exactly as the decision path presents them.
- Do NOT introduce any features that are not in the decision path.
- In the final synthesis, do not state or imply the final classification in the first
  sentence. First explicitly distinguish which conditions SUPPORT the final classification 
  and which conditions OPPOSE it, and explain how the supporting conditions outweigh the
  opposing ones. State the final classification only after this summary.
- Make the explanation coherent, professional, and domain-correct.

Special clarification for checking account status:
- "< 0" means the applicant HAS a checking account, and the account balance is below 0 
  DM (negative balance).

Input:
Applicant features: {features}
Decision path: {decision_path}
Classification: {label}

Output format:
<think>
Step 1: ...
Step 2: ...
...
Overall, [Please state final synthesis], these financial conditions collectively indicate 
that the applicant is classified as "{label}".
</think>
<answer> {label} </answer>
\end{verbatim}
\end{small}

\textbf{Diabetes Dataset}
\begin{small}
\begin{verbatim}
You are a medical reasoning assistant for diabetes monitoring.
Generate a step-by-step rationale that follows exactly the conditions on the given
diagnostic decision rule path.
Do NOT mention terms like "decision tree", "splitting", or "branches".

Instructions:
- Use only the features that appear in the diagnostic path.
- For each step, describe it purely as a factual statement about the patient:
  explicitly state the exact numeric feature value and its relation to the cutoff
  (<= or >) only for continuous numeric features, then explain the clinical and
  metabolic implication of that fact in clear, medically grounded terms related to
  diabetes risk.
- Keep the steps ordered exactly as the diagnostic path presents them.
- Do NOT introduce any features that are not in the diagnostic path.
- In the final synthesis, do not state or imply the final classification in the first
  sentence. First explicitly distinguish which conditions SUPPORT the final
  classification and which conditions OPPOSE it, and explain how the supporting
  conditions outweigh the opposing ones. State the final classification only after 
  this summary.
- Make the explanation clinically coherent, professional, and medically accurate.

Input:
Patient features: {features}
Diagnostic rule path: {decision_path}
Diagnosis: {label}

Output format:
<think>
Step 1: ...
Step 2: ...
...
Overall, [Please state final synthesis], these physiological indicators collectively
indicate that the patient is diagnosed as "{label}".
</think>
<answer> {label} </answer>

\end{verbatim}
\end{small}

\textbf{Homeloan Dataset}
\begin{small}
\begin{verbatim}
You are a home loan approval reasoning assistant.
Your task is to generate a clear, step-by-step rationale that follows exactly the
conditions on the given decision rule path.
Do NOT mention terms like "decision tree", "splitting", or "branches".

Instructions:
- Use only the features that appear in the decision path.
- For each step, describe it purely as a factual statement about the applicant:
  explicitly state the exact numeric feature value and its relation to the cutoff
  (<= or >) only for continuous numeric features, but for categorical features
  state only the real-world category without mentioning any numeric values,
  indicators, encodings, or cutoffs; then explain the implication in clear
  financial and credit-risk terms.
- Keep the steps ordered exactly as the decision path presents them.
- Do NOT introduce any features that are not in the decision path.
- In the final synthesis, do not state or imply the final classification in the
  first sentence. First explicitly distinguish which conditions SUPPORT the final
  classification and which conditions OPPOSE it, and explain how the supporting
  conditions outweigh the opposing ones. State the final classification only after 
  this summary.
- Make the explanation coherent, professional, and domain-correct.

Input:
Applicant features: {features}
Decision path: {decision_path}
Classification: {label}

Output format:
<think>
Step 1: ...
Step 2: ...
...
Overall, [Please state final synthesis], these characteristics collectively support
a "{label}" decision regarding the home loan application.
</think>
<answer> {label} </answer>
\end{verbatim}
\end{small}

\textbf{Credit-D Dataset}
\begin{small}
\begin{verbatim}
You are a financial distress prediction assistant.
Your task is to generate a clear, step-by-step rationale that follows exactly the 
conditions on the given decision rule path.
Do NOT mention terms like “decision tree”, “splitting”, or “branches”.

Instructions:
- Use only the features that appear in the decision path.
- If a feature appears multiple times in the decision path:
    - You MUST merge all constraints on that feature into a single combined interval 
      statement.
    - You MUST present it at the very first occurrence of that feature in the path.
    - Do NOT describe the same feature in multiple separate steps.
- For each step, describe it purely as a factual statement about the applicant—for 
  continuous numeric features, explicitly state the exact value and whether it is <= 
  or > the cutoff—and then explain what this condition implies for the applicant's 
  financial distress risk.
- Keep the steps ordered exactly as the decision path presents them.
- Do NOT introduce any features that are not in the decision path.
- In the final synthesis, do not state or imply the final classification in the first 
  sentence.
  First explicitly distinguish which conditions SUPPORT the final classification and 
  which conditions OPPOSE it and explain how the supporting conditions outweigh the 
  opposing ones.
  State the final classification only after this summary.
- Make the explanation coherent, professional, and domain-correct.

Applicant features: {features}
Decision path: {decision_path}
classification: {label}

Output format:
<think>
Step 1: ...
Step 2: ...
...
Overall, [Please state final synthesis], these financial conditions collectively indicate 
that the applicant is classified as "{label}".
</think>
<answer> {label} </answer>
\end{verbatim}
\end{small}

\subsubsection{Supervised Fine-tuning}
\label{appendix:SFT_scaofflod_reasoning}
We adopt a chat-style prompting format for supervised fine-tuning (SFT), where each training example is explicitly structured into three roles: \textit{system}, \textit{user}, and \textit{assistant}.
The \textit{system} message specifies global behavioral constraints, the \textit{user} message provides the serialized tabular input and task instructions, and the \textit{assistant} message contains the target output with intermediate reasoning.

We use the tokenizer’s \texttt{apply\_chat\_template} function to convert the multi-role chat structure into a single token sequence.
During SFT, loss is computed only on tokens corresponding to the \textit{assistant} role, while tokens from the \textit{system} and \textit{user} roles are masked out.

\textbf{Alzheimer's Diasease Dataset}
\begin{small}
\begin{verbatim}
System:
You are a medical reasoning assistant for Alzheimer's monitoring.  
Given patient information, generate a clinically interpretable step-by-step rationale 
that explains the diagnosis.  
Conclude with a clear diagnostic label ("Alzheimer's Disease", "Mild Cognitive Impairment", 
or "Normal Cognition").  

Guidelines:  
- Present reasoning in a logical step-by-step sequence.  
- End with a concise synthesis that links the findings to the diagnosis.  
- In the final synthesis, do not state or imply the final classification in the first 
  sentence.
  * First summarize the evidence balance:
    - List which steps SUPPORT each class
      (Normal Cognition / Mild Cognitive Impairment / Alzheimer's Disease).
    - Some steps may be neutral and support no class; this is acceptable.
    - Explain how the distribution and strength of supporting evidence across steps leads 
      to one class outweighing the others.
    - Importantly, the final conclusion must be based solely on the cl1inical meaning of 
      the evidence described in each step.
  * Only AFTER that, state the final classification as exactly one of the allowed classes.
- The final answer MUST strictly follow the format below.

Output format:  
<think>  
Step 1: ...  
Step 2: ...  
...
Overall, ...
</think>  
<answer> Alzheimer's Disease / Mild Cognitive Impairment / Normal Cognition </answer>

User:
Here are the patient's features: {patient_features}. Please analyze them and provide a 
diagnostic reasoning and final label.

Assitant: 
{Intermediate Reasoning and Answer}
\end{verbatim}
\end{small}
\textbf{Creditg Dataset}
\begin{small}
\begin{verbatim}
System:
You are a financial risk reasoning assistant.
Given applicant information, generate a clear, step-by-step rationale that explains 
the credit risk classification.
Conclude with a final label ("good" or "bad").

Guidelines:
- Present reasoning in a logical, step-by-step sequence.
- End with a concise synthesis that links the findings to the decision. 
- In the final synthesis, do not state or imply the final classification in the first 
sentence.
  First explicitly distinguish which conditions SUPPORT the final classification and 
  which conditions OPPOSE it and explain how the supporting conditions outweigh the 
  opposing ones.
  State the final classification only after this summary.
- The final answer MUST strictly follow the format below.

Output format:
<think>
Step 1: ...
Step 2: ...
...
Overall, ...
</think>
<answer> good / bad </answer>

User:
Applicant information: {patient_features}. Please generate a step-by-step financial 
reasoning based on these features and provide the final credit risk label ('good' or 'bad').

Assitant: 
{Intermediate Reasoning and Answer}
\end{verbatim}
\end{small}
\textbf{Diabetes Dataset}
\begin{small}
\begin{verbatim}
System:
You are a medical reasoning assistant for diabetes diagnosis.
Given patient information,  generate a clear, step-by-step clinical rationale that explains 
the diabetes classification.
Conclude with a final label ("diabetic" or "non-diabetic").

Guidelines:
- Present reasoning in a logical, step-by-step sequence.
- End with a concise synthesis that links the findings to the decision. 
- In the final synthesis, do not state or imply the final classification in the first 
  sentence.
  First explicitly distinguish which conditions SUPPORT the final classification and which 
  conditions OPPOSE it and explain how the supporting conditions outweigh the opposing ones.
  State the final classification only after this summary.
- The final answer MUST strictly follow the format below.

Output format:
<think>
Step 1: ...
Step 2: ...
...
Overall, ...
</think>
<answer> diabetic / non-diabetic </answer>

User:
Here are the patient's features: {patient_features}. Please analyze them and provide a 
diagnostic reasoning and final label.

Assitant: 
{Intermediate Reasoning and Answer}
\end{verbatim}
\end{small}
\textbf{Homeloan Dataset}
\begin{small}
\begin{verbatim}
System:
You are a home loan approval reasoning assistant.
Given applicant financial and credit information, generate a clear, step-by-step 
rationale that explains the home loan approval decision.
Conclude with a final label ("Approved" or "Not Approved").

Guidelines:
- Present reasoning in a logical step-by-step sequence.  
- End with a concise synthesis that links the findings to the decision.  
- In the final synthesis, do not state or imply the final classification in the first 
  sentence.
  First explicitly distinguish which conditions SUPPORT the final classification and 
  which conditions OPPOSE it and explain how the supporting conditions outweigh the 
  opposing ones.
  State the final classification only after this summary.
- The final answer MUST strictly follow the format below.

Output format:
<think>
Step 1: ...
Step 2: ...
...
Overall, ...
</think>
<answer> Approved / Not Approved </answer>

User:
Here are the applicant's features: {applicant_features}. Please analyze them and provide 
a home loan approval reasoning and final decision.

Assitant: 
{Intermediate Reasoning and Answer}
\end{verbatim}
\end{small}

\textbf{Credit-D Dataset}
\begin{small}
\begin{verbatim}
System:
You are a financial distress prediction assistant.
Given applicant information, generate a clear, step-by-step rationale that explains the 
financial distress risk.
Conclude with a final label ("default" or "no_default").

Guidelines:
- Present reasoning in a logical, step-by-step sequence.
- End with a concise synthesis that links the findings to the decision. 
- In the final synthesis, do not state or imply the final classification in the first sentence.
  First explicitly distinguish which conditions SUPPORT the final classification and which
  conditions OPPOSE it and explain how the supporting conditions outweigh the opposing ones.
  State the final classification only after this summary.
- The final answer MUST strictly follow the format below.

Output format:
<think>
Step 1: ...
Step 2: ...
...
Overall, ...
</think>
<answer> default / no_default </answer>

User:
Applicant information: {applicant_features}. Please generate a step-by-step financial distress 
reasoning based on these features and provide the final label."

Assitant: 
{Intermediate Reasoning and Answer}
\end{verbatim}
\end{small}

\subsubsection{Reinforcement Learning}
For reinforcement learning, we use the same format as in supervised fine-tuning. The only difference is that no assistant response is provided in the prompt, as the model is required to generate the reasoning and prediction autonomously during training.
\subsection{Prompts for DRC}
\subsubsection{Reasoning Data Curation}
\label{appendix:prompt_DRC}

\textbf{Alzheimer's Disease Dataset}
\begin{small}
\begin{verbatim}
You are a medical reasoning assistant for Alzheimer's monitoring.  
Your task is to generate a clear, step-by-step rationale expaining the diagnosis of 
Alzheimer's Diasease.

Instructions:  
- Only include the features stated in the patient features.
- You are NOT required to use all features. Select only the features you judge to be the 
  most informative and behaviorally relevant for predicting. 
- Write the rationale as a sequence of steps (Step 1, Step 2, ...), each focusing on ONE 
  feature or a tightly related pair of features.
- For each step:
  * First, state a purely factual clinical description of the patient based on the given 
    condition:
    - For continuous numeric clinical features, explicitly state the exact measured value.
    - For categorical clinical features, state only the real-world clinical category.
    - For ordinal clinical features, explicitly state the true ordinal value first (0, 1, 
      or 2) as observed in the data.
  * Then, explain the clinical implication of this finding in clear Alzheimer's disease-
    related diagnostic terms.
- In the final synthesis, do not state or imply the final classification in the first 
  sentence.
  * First summarize the evidence balance:
    - List which steps SUPPORT each class
      (Normal Cognition / Mild Cognitive Impairment / Alzheimer's Disease).
    - Some steps may be neutral and support no class; this is acceptable.
    - Explain how the distribution and strength of supporting evidence across steps leads 
      to one class outweighing the others.
  * Only AFTER that, state the final classification as exactly one of the allowed classes.
- Make the explanation coherent, professional, and domain-correct.
  
Interpret features:
- Brain regions: 0=normal, 1=mild atrophy, 2=severe atrophy.
- Lateral Ventricle shape: 0=round, 1=moderate change, 2=sharp (abnormal).
- Vascular disease: 0=no, 1=yes.
- MMSE: lower = worse cognition.
- APOE4: 0=no allele, 1=one allele, 2=two alleles (higher risk).
- Education: higher = more cognitive reserve.
- Age: higher age increases Alzheimer's risk; younger age makes significant 
  neurodegeneration less likely.

Input:
Patient data: {patient_features}
Classification: {label}

Output format:
<think>
Step 1: ...
Step 2: ...
...
Overall, [Please state final synthesis], these characteristics collectively support a 
"{label}" classification.
</think>
<answer> {label} </answer>
\end{verbatim}
\end{small}
\textbf{Creditg Dataset}
\begin{small}
\begin{verbatim}
You are a credit risk reasoning assistant.
Your task is to generate a clear, step-by-step rationale explaining whether the applicant 
is a good or bad credit risk.

Instructions:
- Use ONLY the information explicitly stated in the applicant features. Do NOT invent any 
  additional facts.
- You are NOT required to use all features. Select only the features you judge to be the 
  most informative and behaviorally relevant for predicting.
- Write the rationale as a sequence of steps (Step 1, Step 2, ...), each focusing on ONE 
  feature or a tightly related pair of features.
- For each step:
  * First, state a purely factual description of the applicant using the exact value 
    (for numeric features) or the real-world category (for categorical features).
  * Then, explain what this implies in clear credit risk assessment terms.
- In the final synthesis, do not state or imply the final classification in the first 
  sentence. First explicitly distinguish which conditions SUPPORT the final classification 
  and which conditions OPPOSE it and explain how the supporting conditions outweigh the 
  opposing ones. State the final classification only after this summary.
- Make the explanation coherent, professional, and domain-correct.

Input:
Applicant features: {features}
Classification: {label}

Output format:
<think>
Step 1: ...
Step 2: ...
...
Overall, [Please state final synthesis], these financial conditions collectively indicate 
that the applicant is classified as "{label}".
</think>
<answer> {label} </answer>
\end{verbatim}
\end{small}
\textbf{Diabetes Dataset}
\begin{small}
\begin{verbatim}
You are a diabetes risk reasoning assistant.
Your task is to generate a clear, step-by-step rationale explaining whether the individual 
is likely to have diabetes.

Instructions:
- Use ONLY the information explicitly stated in the individual’s clinical and demographic 
  features. Do NOT invent any additional facts.
- You are NOT required to use all features. Select only the features you judge to be the 
  most informative and clinically relevant for diabetes prediction.
- Write the rationale as a sequence of steps (Step 1, Step 2, ...), each focusing on ONE 
  feature or a tightly related pair of features.
- For each step:
  * First, state a purely factual description of the individual using the exact value 
    (for numeric features) or the real-world category (for categorical features).
  * Then, explain what this implies in clear diabetes risk or metabolic health terms.
- In the final synthesis, do not state or imply the final classification in the first 
  sentence. First explicitly distinguish which conditions SUPPORT the final classification 
  and which conditions OPPOSE it and explain how the supporting conditions outweigh the 
  opposing ones. State the final classification only after this summary.
- Make the explanation coherent, professional, and medically domain-correct.

Input:
Individual features: {features}
Classification: {label}

Output format:
<think>
Step 1: ...
Step 2: ...
...
Overall, [Please state final synthesis], these clinical and metabolic conditions 
collectively indicate that the individual is classified as "{label}".
</think>
<answer> {label} </answer>
\end{verbatim}
\end{small}
\textbf{Homeloan Dataset}
\begin{small}
\begin{verbatim}
You are a home loan approval reasoning assistant.
Your task is to generate a clear, step-by-step rationale explaining whether the applicant 
is likely to be Approved or Not Approved for a home loan.

Instructions:
- Use ONLY the information explicitly stated in the applicant features. Do NOT invent any 
  additional facts.
- You are NOT required to use all features. Select only the features you judge to be the 
  most informative and relevant for home loan approval decisions.
- Write the rationale as a sequence of steps (Step 1, Step 2, ...), each focusing on ONE 
  feature or a tightly related pair of features.
- For each step:
  * First, state a purely factual description of the applicant using the exact value 
    (for numeric features) or the real-world category (for categorical features).
  * Then, explain what this implies in clear home loan underwriting, affordability, and 
    repayment-risk terms.
- In the final synthesis, do not state or imply the final classification in the first 
  sentence. First explicitly distinguish which conditions SUPPORT the final classification 
  and which conditions OPPOSE it and explain how the supporting conditions outweigh the 
  opposing ones. State the final classification only after this summary.
- Make the explanation coherent, professional, and domain-correct.

Input:
Applicant features: {features}
Classification: {label}

Output format:
<think>
Step 1: ...
Step 2: ...
...
Overall, [Please state final synthesis], these characteristics collectively support a
"{label}" decision regarding the home loan application.
</think>
<answer> {label} </answer>
\end{verbatim}
\end{small}

\textbf{Credit-D Dataset}
\begin{small}
\begin{verbatim}
You are a financial distress risk analysis assistant.
Your task is to generate a clear, step-by-step rationale explaining whether the individual 
is likely to experience financial distress.

Instructions:
- Use ONLY the information explicitly stated in the individual’s features. Do NOT invent 
  any additional facts.
- You are NOT required to use all features. Select only the features you judge to be the 
  most informative and behaviorally relevant.
- Write the rationale as a sequence of steps (Step 1, Step 2, ...), each focusing on ONE 
  feature or a tightly related pair of features.
- For each step:
  * First, state a purely factual description using the exact value (for numeric features) 
  or the real-world category (for categorical features).
  * Then, explain what this implies in terms of financial stability or distress risk.
- In the final synthesis:
  * Do NOT state or imply the final classification in the first sentence.
  * First explicitly distinguish which conditions SUPPORT financial distress and which 
  conditions OPPOSE it.
  * Then explain how the supporting conditions outweigh the opposing ones.
  * Only after that, state the final classification.
- Make the explanation coherent, professional, and domain-correct.

Input:
Individual features: {features}
Classification: {label}

Output format:
<think>
Step 1: ...
Step 2: ...
...
Overall, [Please state final synthesis], these financial conditions collectively indicate 
that the individual is classified as "{label}".
</think>
<answer> {label} </answer>
\end{verbatim}
\end{small}

\subsubsection{Supervised Fine-tuning}
For supervised fine-tuning, we adopt the same prompting format as used in ReSS. The detailed prompt specification is provided in Appendix~\ref{appendix:SFT_scaofflod_reasoning}.

\subsection{Delexicalized Reasoning Prompts}
\label{appendix:prompt_delexicalized}
\subsubsection{Delexicalized Reasoning Data Curation}
Under this setting, we adopt a single unified prompt to generate reasoning data across all datasets. 
\begin{small}
\begin{verbatim}
You are a decision rule execution assistant.
Your task is to generate a step-by-step rationale that follows **exactly the conditions 
on the given decision rule path**.
This is NOT an explanation task.
Do NOT interpret, explain, or justify the conditions in domain or behavioral terms.
Do NOT mention terms like “decision tree”, “splitting”, or “branches”.

Instructions:
- Use ONLY the features that appear in the decision path.
- Follow the decision path strictly and in the given order.
- For each step:
  * State a purely factual condition check, without interpretation.
  * For continuous numeric features and ordinal features:
    - Explicitly state the exact numeric value and whether it is <= or > the cutoff.
  * For categorical (one-hot encoded) features:
    - Treat "> 0.500" as **the instance belongs to the specified category**.
    - Treat "<= 0.500" as **the instance does NOT belong to the specified category**.
    - Explicitly state whether the instance belongs to the category or not.
    - **Do NOT mention any numeric thresholds (e.g., 0.500), indicator values, encodings, 
    or one-hot representations.**
  * Do NOT explain why the condition matters.
- Do NOT introduce any features that are not in the decision path.
- Do NOT add behavioral, causal, or domain-level explanations.
- The final decision must be derived strictly from executing all conditions on the path.

Input:
Features: {features}
Decision path: {decision_path}

Output format:
<think>
Step 1:
Step 2:
...
Therefore, the final decision is "{label}".
</think>
<answer> {label} </answer>   
\end{verbatim}
\end{small}
\label{appendix:prompt_wo_symbolic_scaffold_generation}

\subsubsection{Supervised Fine-Tuning }
\label{appendix:SFT_delexicalized_reasoning}
For supervised fine-tuning on delexicalized reasoning data, we adopt a single unified prompt across all datasets. each training example is explicitly structured into three roles: \textit{system}, \textit{user}, and \textit{assistant}.
\begin{small}
\begin{verbatim}
System:
You are a decision-making assistant.

Given the input features, generate a clear, step-by-step rationale and the final decision.

Guidelines:
- Use only the information provided in the input.
- Do NOT introduce any domain-specific background, assumptions, or explanations.
- The final answer MUST strictly follow the format below.

Output format:
<think>
Step 1: ...
Step 2: ...
...
Therefore, ...
</think>
<answer> Label </answer>

User:
Here is the input features: {input_features}. Please generate a clear, step-by-step 
rationale and the final decision.

Assistant:
{Delexicalized Reasoning and Answer}
\end{verbatim}
\end{small}
\subsection{Prompts for Direct SFT without Intermediate Reasoning}
\label{appendix:direct_SFT}
We use the same chat-style prompting framework as in the reasoning-based setting, with each example organized into \textit{system}, \textit{user}, and \textit{assistant} roles.
The \textit{system} and \textit{user} messages define the task and provide the serialized tabular input, while the \textit{assistant} message contains only the final prediction, without any intermediate reasoning.

The multi-role prompt is linearized using the tokenizer’s \texttt{apply\_chat\_template} function.
During supervised fine-tuning, loss is computed exclusively on tokens from the \textit{assistant} role, with all \textit{system} and \textit{user} tokens masked.
This setup enables a controlled comparison with the reasoning-based model, isolating the effect of intermediate reasoning on predictive performance.

\textbf{Alzheimer's Disease Dataset}
\begin{small}
\begin{verbatim}
System:
A conversation between User and Assistant. The user asks a question, and the Assistant 
solves it. The answer is enclosed within <answer> </answer> tag, i.e.,
<answer> answer here </answer>.

User:
What is the diagnosis of the patient based on the patient information: {patient_feature}?
Answer me with only either 'Alzheimer's Disease', 'Mild Cognitive Impairment', 
or 'Normal Cognition'.

Assitant: 
{Prediction Answer Only}
\end{verbatim}
\end{small}
\textbf{Creditg Dataset}
\begin{small}
\begin{verbatim}
System:
A conversation between User and Assistant. The user asks a question, and the Assistant 
solves it. The answer is enclosed within <answer> </answer> tag, i.e.,
<answer> answer here </answer>.

User:
What is the credit risk classification of the applicant based on the information: 
{patient_features}? Answer me with only either 'good' or 'bad'.

Assitant: 
{Prediction Answer Only}
\end{verbatim}
\end{small}
\textbf{Diabetes Dataset}
\begin{small}
\begin{verbatim}
System:
A conversation between User and Assistant. The user asks a question, and the Assistant 
solves it. The answer is enclosed within <answer> </answer> tag, i.e.,
<answer> answer here </answer>.

User:
What is the diagnosis of the patient based on the patient information: {patient_features}?
Answer me with only either 'diabetic' or 'non-diabetic'. 

Assitant: 
{Prediction Answer Only}
\end{verbatim}
\end{small}
\textbf{Homeloan Dataset}
\begin{small}
\begin{verbatim}
System:
A conversation between User and Assistant. The user asks a question, and the Assistant 
solves it. The answer is enclosed within <answer> </answer> tag, i.e.,
<answer> answer here </answer>.

User:
What is the home loan approval decision based on the applicant information: 
{applicant_features}? Answer with only either 'Approved' or 'Not Approved'. 

Assitant: 
{Prediction Answer Only}
\end{verbatim}
\end{small}

\textbf{Credit-D Dataset}
\begin{small}
\begin{verbatim}
System:
A conversation between User and Assistant. The user asks a question, and the Assistant 
solves it. The answer is enclosed within <answer> </answer> tag, i.e.,
<answer> answer here </answer>.

User:
"Will the borrower experience serious financial distress in the next two years based on 
the information: {applicant_features}?". Answer with only either 'default' or 'non-default'." 

Assitant: 
{Prediction Answer Only}
\end{verbatim}
\end{small}

\subsection{Prompts for GPT-4.1 with Intermediate Reasoning}
\label{appendix:prompt_GPT_reasoning}
We invoke the GPT-4.1 model via the OpenAI API and explicitly prompt it to generate intermediate reasoning before producing the final answer.
Each prompt is explicitly structured into two components: a \textit{System} message, which defines the model's role and global behavioral constraints, and a \textit{User} message, which provides the task-specific instructions and input instance.

All generations are performed with deterministic decoding, using the following parameters:
\texttt{temperature}=0, \texttt{top\_p}=1, and \texttt{max\_tokens}=1024.

\textbf{Alzheimer's Disease Dataset}
\begin{small}
\begin{verbatim}
System:
You are a helpful and precise reasoning assistant.

User:
You are a medical reasoning assistant for Alzheimer's monitoring.  
Given patient information, generate a clinically interpretable step-by-step rationale that 
explains the diagnosis.  
Conclude with a clear diagnostic label ("Alzheimer's Disease", "Mild Cognitive Impairment", 
or "Normal Cognition").  

Guidelines:  
- Present reasoning in a logical step-by-step sequence.  
- End with a concise synthesis that links the findings to the diagnosis.  
- In the final synthesis, do not state or imply the final classification in the first 
sentence.
  * First summarize the evidence balance:
    - List which steps SUPPORT each class
      (Normal Cognition / Mild Cognitive Impairment / Alzheimer's Disease).
    - Some steps may be neutral and support no class; this is acceptable.
    - Explain how the distribution and strength of supporting evidence across steps leads 
      to one class outweighing the others.
    - Importantly, the final conclusion must be based solely on the
      clinical meaning of the evidence described in each step.
  * Only AFTER that, state the final classification as exactly one of the allowed classes.
- The final answer MUST strictly follow the format below.

Input:
Patient data: {case_features}

Output format:  
<think>  
Step 1: ...  
Step 2: ...  
...
Overall, ...
</think>  
<answer> Alzheimer's Disease / Mild Cognitive Impairment / Normal Cognition </answer>
\end{verbatim}
\end{small}

\textbf{Creditg Dataset}
\begin{small}
\begin{verbatim}
System:
You are a helpful and precise reasoning assistant.

User:
You are a financial risk reasoning assistant.
Given applicant information, generate a clear, step-by-step rationale that explains the 
credit risk classification.
Conclude with a final label ("good" or "bad").

Guidelines:
- Present reasoning in a logical, step-by-step sequence.
- End with a concise synthesis that links the findings to the decision. 
- In the final synthesis, do not state or imply the final classification in the first 
  sentence.
  First explicitly distinguish which conditions SUPPORT the final classification and which 
  conditions OPPOSE it and explain how the supporting conditions outweigh the opposing ones.
  State the final classification only after this summary.
- The final answer MUST strictly follow the format below.

Input:
Applicant features: {case_features}

Output format:
<think>
Step 1: ...
Step 2: ...
...
Overall, ...
</think>
<answer> good / bad </answer>
\end{verbatim}
\end{small}
\textbf{Diabetes Dataset}
\begin{small}
\begin{verbatim}
System:
You are a helpful and precise reasoning assistant.

User:
You are a medical reasoning assistant for diabetes diagnosis.
Given patient information,  generate a clear, step-by-step clinical rationale that explains 
the diabetes classification.
Conclude with a final label ("diabetic" or "non-diabetic").

Guidelines:
- Present reasoning in a logical, step-by-step sequence.
- End with a concise synthesis that links the findings to the decision. 
- In the final synthesis, do not state or imply the final classification in the first 
  sentence.
  First explicitly distinguish which conditions SUPPORT the final classification and which 
  conditions OPPOSE it and explain how the supporting conditions outweigh the opposing ones.
  State the final classification only after this summary.
- The final answer MUST strictly follow the format below.

Input:
Individual features: {case_features}

Output format:
<think>
Step 1: ...
Step 2: ...
...
Overall, ...
</think>
<answer> diabetic / non-diabetic </answer>
\end{verbatim}
\end{small}
\textbf{Homeloan Dataset}
\begin{small}
\begin{verbatim}
System:
You are a helpful and precise reasoning assistant.

User:
You are a home loan approval reasoning assistant.
Given applicant financial and credit information, generate a clear, step-by-step rationale 
that explains the home loan approval decision.
Conclude with a final label ("Approved" or "Not Approved").

Guidelines:
- Present reasoning in a logical step-by-step sequence.  
- End with a concise synthesis that links the findings to the decision.  
- In the final synthesis, do not state or imply the final classification in the first 
  sentence.
  First explicitly distinguish which conditions SUPPORT the final classification and which 
  conditions OPPOSE it and explain how the supporting conditions outweigh the opposing ones.
  State the final classification only after this summary.
- The final answer MUST strictly follow the format below.

Input:
Applicant features: {case_features}

Output format:
<think>
Step 1: ...
Step 2: ...
...
Overall, ...
</think>
<answer> Approved / Not Approved </answer>
\end{verbatim}
\end{small}

\textbf{Credit-D Dataset}
\begin{small}
\begin{verbatim}
System:
You are a helpful and precise reasoning assistant.

User:
You are a financial distress prediction assistant.
Given applicant information, generate a clear, step-by-step rationale that explains the 
financial distress risk.
Conclude with a final label ("default" or "no_default").

Guidelines:
- Present reasoning in a logical, step-by-step sequence.
- End with a concise synthesis that links the findings to the decision. 
- In the final synthesis, do not state or imply the final classification in the first sentence.
  First explicitly distinguish which 
  conditions SUPPORT the final classification and which conditions OPPOSE it and explain how 
  the supporting conditions outweigh the opposing ones.
  State the final classification only after this summary.
- The final answer MUST strictly follow the format below.

Input:
Applicant feartures: {case_features}

Output format:
<think>
Step 1: ...
Step 2: ...
...
Overall, ...
</think>
<answer> default / no_default </answer>
\end{verbatim}
\end{small}

\subsection{Prompts for GPT-4.1 without Intermediate Reasoning}
\label{appendix:prompt_GPT_wo_reasoning}
We invoke the GPT-4.1 model via the OpenAI API to generate predictions without explicitly requesting intermediate reasoning.
Each prompt is structured into two components: a \textit{System} message, which specifies the assistant's role and global behavioral constraints, and a \textit{User} message, which provides the task description and input features, and instructs the model to directly produce the final answer.

All generations are performed with deterministic decoding, using the same hyperparameters as in the reasoning-based setting:
\texttt{temperature}=0, \texttt{top\_p}=1.

\textbf{Alzheimer's Disease Dataset}
\begin{small}
\begin{verbatim}
Syetem:
A conversation between User and Assistant. The user asks a question, and the Assistant 
solves it. The answer is enclosed within <answer> </answer> tag, i.e.,
<answer> answer here </answer>.

User:
What is the diagnosis of the patient based on the patient information: {case_features}? 
Answer me with only either 'Alzheimer's Disease', 'Mild Cognitive Impairment', 
or 'Normal Cognition'. The final answer must be enclosed within <answer></answer> tags, 
i.e., <answer> answer here </answer>.
\end{verbatim}
\end{small}
\textbf{Creditg Dataset}
\begin{small}
\begin{verbatim}
Syetem:
A conversation between User and Assistant. The user asks a question, and the Assistant 
solves it. The answer is enclosed within <answer> </answer> tag, i.e.,
<answer> answer here </answer>.

User:
What is the credit risk classification of the applicant based on the information: 
{case_features}? 
Answer me with only either 'good' or 'bad'. The final classification must be enclosed 
within <answer> </answer> tags, i.e., <answer> answer here </answer>.    
\end{verbatim}
\end{small}
\textbf{Diabetes Dataset}
\begin{small}
\begin{verbatim}
Syetem:
A conversation between User and Assistant. The user asks a question, and the Assistant 
solves it. The answer is enclosed within <answer> </answer> tag, i.e.,
<answer> answer here </answer>. 

User:
What is the diabetes diagnosis of the patient based on the following information: 
{case_features}?
Answer me with only either 'non-diabetic' or 'diabetic'. The final answer must be enclosed 
within <answer></answer> tags, i.e., <answer> answer here </answer>.
\end{verbatim}
\end{small}
\textbf{Homeloan Dataset}
\begin{small}
\begin{verbatim}
Syetem:
A conversation between User and Assistant. The user asks a question, and the Assistant 
solves it. The answer is enclosed within <answer> </answer> tag, i.e.,
<answer> answer here </answer>.

User:
What is the home loan approval decision based on the applicant information: {case_features}? 
Answer with only either 'Approved' or 'Not Approved'. The answer must be enclosed within 
<answer> </answer> tags, i.e., <answer> answer here </answer>.
\end{verbatim}
\end{small}

\textbf{Credit-D Dataset}
\begin{small}
\begin{verbatim}
Syetem:
A conversation between User and Assistant. The user asks a question, and the Assistant 
solves it. The answer is enclosed within <answer> </answer> tag, i.e.,
<answer> answer here </answer>.

User:
Will the borrower experience serious financial distress in the next two years based on 
the information: {case_features}? Answer me with only either 'default' or 'non-default'.
The answer must be enclosed within <answer> </answer> tags, i.e., <answer> answer here </answer>.
\end{verbatim}
\end{small}
\newpage

\subsection{Prompts for GPT-4.1 with Scaffold-Augmented Reasoning}
\label{appendix:prompt_GPT_w_scaffold_reasoning}
We invoke the GPT-4.1 model via the OpenAI API to generate predictions with intermediate reasoning, where the symbolic decision path $S(\mathbf{x})$ extracted from the trained decision tree is additionally provided as part of the input prompt to guide the reasoning process.
Each prompt is structured into two components: a \textit{System} message, which specifies the assistant's role and global behavioral constraints, and a \textit{User} message, which provides the task description, input features, and the corresponding symbolic decision path, and instructs the model to generate step-by-step reasoning before producing the final answer.

All generations are performed with deterministic decoding, using the same hyperparameters as in the reasoning-based setting:
\texttt{temperature}=0, \texttt{top\_p}=1.

\textbf{Alzheimer's Disease Dataset}
\begin{small}
\begin{verbatim}
Syetem:
You are a helpful and precise reasoning assistant.

User:
You are a medical reasoning assistant for Alzheimer's monitoring.

Your task is to generate a clear, step-by-step rationale for the diagnosis by STRICTLY 
following the provided decision path, and then infer the final diagnosis.

Instructions:
- You are given:
  1. Patient features
  2. A decision path produced by a decision tree
- Follow the decision path EXACTLY in order.
- Only use features that appear in the decision path.
- Do NOT introduce additional features, outside evidence, or new reasoning steps beyond 
  the path.
- Each step must correspond to one condition in the decision path.
- For each step:
  * First, state a purely factual clinical description based on the patient data and the 
    path condition:
    - For continuous numeric clinical features, explicitly state the exact measured value
      and how it compares with the threshold in the path.
    - For categorical clinical features, state the real-world clinical category and how it
      matches the path condition.
    - For ordinal clinical features, explicitly state the true ordinal value first (0, 1, or 2), 
      and then relate it to the path condition.
  * Then, explain the clinical implication of this finding in clear Alzheimer's disease-related 
    diagnostic terms.
- The final diagnosis must be inferred ONLY from the evidence described in the steps above.
- In the final synthesis, do not state or imply the final classification in the first sentence.
  * First summarize the evidence balance:
    - List which steps SUPPORT each class
      (Normal Cognition / Mild Cognitive Impairment / Alzheimer's Disease).
    - Some steps may be neutral and support no class; this is acceptable.
    - Explain how the distribution and strength of supporting evidence across steps leads to one 
      class outweighing the others.
  * Only AFTER that, state the final classification as exactly one of the allowed classes.
- Make the explanation coherent, professional, and faithful to the provided path.

Interpret features:
- Brain regions: 0=normal, 1=mild atrophy, 2=severe atrophy.
- Lateral Ventricle shape: 0=round, 1=moderate change, 2=sharp (abnormal).
- MMSE: lower = worse cognition.
- APOE4: 0=no allele, 1=one allele, 2=two alleles (higher risk).
- Education: higher = more cognitive reserve.
- Age: higher age increases Alzheimer's risk; younger age makes significant neurodegeneration
  less likely.

Input:
Patient data:
{case_features}

Decision path:
{decision_path}

Output format:
<think>
Step 1: ...
Step 2: ...
...
Overall, ...
</think>
<answer> Alzheimer's Disease / Mild Cognitive Impairment / Normal Cognition </answer>
\end{verbatim}
\end{small}

\textbf{Creditg Dataset}
\begin{small}
\begin{verbatim}
Syetem:
You are a helpful and precise reasoning assistant.

User:
You are a credit risk reasoning assistant.

Your task is to generate a clear, step-by-step rationale by STRICTLY following the 
given decision path, and then determine whether the applicant is a good or bad credit risk.

Instructions:
- You are given:
  1. Applicant features
  2. A decision path
- Follow the decision path EXACTLY in order.
- Use ONLY the features that appear in the decision path.
- Do NOT introduce any additional features, outside facts, or new reasoning beyond the path.
- Each step must correspond to one condition in the decision path.
- For each step:
  * First, state a purely factual description of the applicant based on the applicant features 
    and the path condition.
  * For numeric features:
    - explicitly state the exact value
    - explicitly state whether it is <= or > the cutoff in the path
  * For categorical features:
    - state only the real-world category
    - do NOT mention numeric encodings, indicators, or hidden values
  * Then, explain what this condition implies in clear credit risk assessment terms.
- The final classification must be inferred ONLY from the reasoning above.

Final synthesis:
- Do NOT state or imply the final classification in the first sentence.
- Explicitly distinguish which steps SUPPORT a "good" credit assessment and which steps 
  SUPPORT a "bad" credit assessment.
- Explain how the balance of evidence leads to one class outweighing the other.
- State the final classification ONLY at the end.

Make the explanation coherent, professional, and faithful to the provided path.

Input:
Applicant features:
{case_features}

Decision path:
{decision_path}

Output format:
<think>
Step 1: ...
Step 2: ...
...
Overall, ...
</think>
<answer>good / bad</answer>
\end{verbatim}
\end{small}

\textbf{Diabetes Dataset}
\begin{small}
\begin{verbatim}
Syetem:
You are a helpful and precise reasoning assistant.

User:
You are a diabetes risk reasoning assistant.

Your task is to generate a clear, step-by-step rationale by STRICTLY following the given 
decision path, and then determine whether the individual is diabetic or non-diabetic.

Instructions:
- You are given:
  1. Individual features
  2. A decision path
- Follow the decision path EXACTLY in order.
- Use ONLY the features that appear in the decision path.
- Do NOT introduce any additional features, outside evidence, or new reasoning beyond the path.
- Each step must correspond to one condition in the decision path.
- For each step:
  * First, state a purely factual description of the individual based on the feature value.
    - For numeric features: explicitly state the exact value and whether it is <= or > the 
      cutoff in the path.
    - For categorical features: state the real-world category only (do not mention encodings).
  * Then, explain what this implies in clear diabetes risk or metabolic health terms.
- The final classification must be inferred ONLY from the reasoning above.

Final synthesis:
- Do NOT state or imply the final classification in the first sentence.
- First distinguish:
  * which steps SUPPORT a diabetic outcome
  * which steps SUPPORT a non-diabetic outcome
- Explain how the balance of evidence leads to one outcome outweighing the other.
- Only AFTER that, state the final classification.

Make the explanation coherent, professional, and faithful to the provided path.

Input:
Individual features:
{case_features}

Decision path:
{decision_path}

Output format:
<think>
Step 1: ...
Step 2: ...
...
Overall, ...
</think>
<answer>diabetic / non-diabetic</answer>
\end{verbatim}
\end{small}

\textbf{Homeloan Dataset}
\begin{small}
\begin{verbatim}
Syetem:
You are a helpful and precise reasoning assistant.

User:
You are a credit risk reasoning assistant.

Your task is to generate a clear, step-by-step rationale by STRICTLY following the given 
decision path, and then determine whether the applicant is Approved or Not Approved for a
home loan.

Instructions:
- You are given:
  1. Applicant features
  2. A decision path
- Follow the decision path EXACTLY in order.
- Use ONLY the features that appear in the decision path.
- Do NOT introduce any additional features, outside facts, or new reasoning beyond the path.
- Each step must correspond to one condition in the decision path.
- For each step:
  * First, state a purely factual description of the applicant based on the applicant features 
  and the path condition.
  * For numeric features:
    - explicitly state the exact value
    - explicitly state whether it is <= or > the cutoff in the path
  * For categorical features:
    - state only the real-world category
    - do NOT mention numeric encodings, indicators, or hidden values
  * Then, explain what this condition implies in clear home loan underwriting, affordability, 
    and repayment-risk terms.
- The final classification must be inferred ONLY from the reasoning above.

Final synthesis:
- Do NOT state or imply the final classification in the first sentence.
- First explicitly distinguish:
  * which steps SUPPORT an Approved decision
  * which steps SUPPORT a Not Approved decision
- Explain how the balance of evidence leads to one outcome outweighing the other.
- Only AFTER that, state the final classification.

Make the explanation coherent, professional, and faithful to the provided path.

Input:
Applicant features:
{case_features}

Decision path:
{decision_path}

Output format:
<think>
Step 1: ...
Step 2: ...
...
Overall, ...
</think>
<answer> Approved / Not Approved </answer>
\end{verbatim}
\end{small}

\textbf{Credit-D Dataset}
\begin{small}
\begin{verbatim}
Syetem:
You are a helpful and precise reasoning assistant.

User:
You are a credit risk reasoning assistant.

Your task is to generate a clear, step-by-step rationale by STRICTLY following the given 
decision path, and then determine whether the borrower will default or no_default.

Instructions:
- You are given:
  1. Borrower features
  2. A decision path
- Follow the decision path EXACTLY in order.
- Use ONLY the features that appear in the decision path.
- Do NOT introduce any additional features, outside facts, or new reasoning beyond the path.
- Each step must correspond to one condition in the decision path.
- For each step:
  * First, state a purely factual description of the borrower based on the borrower features 
    and the path condition.
  * For numeric features:
    - explicitly state the exact value
    - explicitly state whether it is <= or > the cutoff in the path
  * For categorical features:
    - state only the real-world category
    - do NOT mention numeric encodings, indicators, or hidden values
  * Then, explain what this condition implies in clear consumer credit, debt-burden, liquidity, 
    and delinquency-risk terms (e.g., revolving utilization, debt-to-income, past due history, 
    number of credit lines, dependents).
- The final classification must be inferred ONLY from the reasoning above.

Final synthesis:
- Do NOT state or imply the final classification in the first sentence.
- First explicitly distinguish:
  * which steps SUPPORT a no_default decision
  * which steps SUPPORT a default decision
- Explain how the balance of evidence leads to one outcome outweighing the other.
- Only AFTER that, state the final classification.

Make the explanation coherent, professional, and faithful to the provided path.

Input:
Applicant data:
{case_features}

Decision path:
{decision_path}

Output format:
<think>
Step 1: ...
Step 2: ...
...
Overall, ...
</think>
<answer> default / no_default </answer>
\end{verbatim}
\end{small}

\newpage
\section{Reasoning Data Examples}
\label{appendix:reasoning_example}
\subsection{Examples of Reasoning Data Curated by ReSS}
\label{appendix:reasoning_example_ReSS}
\textbf{Alzheimer's Disease Dataset}
\begin{figure}[h]
\centering
\fbox{
\begin{minipage}{0.95\linewidth}

\small
\textbf{Original Features:}
\vspace{0.5em}

The atrophy level of hippocampal is 0. The atrophy level of Amygdala is 0. The atrophy level of entorhinal is 0. The atrophy level of parahippocampal is 0. The atrophy level of medial temporal lobe is 0. The atrophy level of fusiform is 2. The atrophy level of precuneus is 0. The atrophy level of superior parietal is 0. The atrophy level of medial temporal lobe (vs cortex) is 2. The atrophy level of parietal lobe (vs cortex) is 0. The atrophy level of frontal lobe is 0. The atrophy level of temporal lobe is 0. The atrophy level of parietal lobe is 0. The atrophy level of occipital lobe is 0. The overall cortical atrophy is 0. The level of ventricle enlargement is 0. The temporal shape of the lateral ventricle is 2. The frontal shape of the lateral ventricle is 2. The evidence of vascular disease is 0. The age is 68.4 years old. The sex is female. The MMSE score is 27.0. The APOE~$\varepsilon4$ status is 1. The marital status is divorced. The years of education are 14.
\vspace{0.5em}

\textbf{Curated Reasoning by ReSS:}

\vspace{0.5em}
\texttt{<think>}

Step 1: The patient’s MMSE score is 27.0, which is greater than 25.5. This indicates relatively preserved global cognitive function, as MMSE scores above 25 are generally considered within the normal range. In the context of Alzheimer's disease, this finding suggests that there is no significant cognitive impairment at present.

\vspace{0.5em}

Step 2: The patient’s MMSE score is 27.0, which is also less than or equal to 28.5. This further confirms that the patient’s cognitive performance is within the upper-normal to mildly reduced range, but still not in the range typically associated with mild cognitive impairment or dementia. This supports the absence of significant cognitive decline.

\vspace{0.5em}

Step 3: The atrophy level of the hippocampus is 0, which is interpreted as normal. The hippocampus is a key region affected early in Alzheimer’s disease, and normal hippocampal volume strongly argues against both mild cognitive impairment and Alzheimer’s disease.

\vspace{0.5em}

Step 4: The patient’s age is 68.4 years, which is greater than 64.75 years. Being in this age group does increase the baseline risk for Alzheimer’s disease, as advancing age is a major risk factor. However, age alone does not indicate the presence of disease; it only raises the background risk.

\vspace{0.5em}

Step 5: The atrophy level of the temporal lobe is 0, which is interpreted as normal. The temporal lobe, particularly the medial temporal structures, is often affected in Alzheimer’s disease. Normal temporal lobe structure further supports the absence of neurodegeneration typically seen in Alzheimer’s disease.

\vspace{0.5em}

Overall, the evidence supporting Normal Cognition includes the preserved MMSE score (steps 1 and 2), normal hippocampal structure (step 3), and normal temporal lobe structure (step 5). Step 4 (age) increases risk but does not directly indicate disease and is thus neutral. There is no evidence from these steps supporting Mild Cognitive Impairment or Alzheimer’s Disease, as there are no findings of cognitive impairment or neurodegeneration in the key regions assessed. The balance of evidence, therefore, strongly favors a classification of Normal Cognition.

\texttt{</think>}

\texttt{<answer>} Normal Cognition \texttt{</answer>}

\vspace{0.5em}
\end{minipage}
}
\caption{An example of step-by-step reasoning curated by ReSS on AD dataset.}
\end{figure}

\newpage
\textbf{Creditg Dataset}

\begin{figure}[h]
\centering
\fbox{
\begin{minipage}{0.95\linewidth}

\small
\textbf{Original Features:}
\vspace{0.5em}

The Status of existing checking account is no checking account. The Duration in month is 24. The Credit history is existing credits paid back duly till now. The Purpose is furniture/equipment. The Credit amount is 2835. The Savings account/bonds is $100 \leq \cdots < 500$ DM. The Present employment since is $\geq 7$. The Installment rate in percentage of disposable income is 3. The Personal status and sex is male : single. The Other debtors / guarantors is none. The Present residence since is 4. The Property is building society savings agreement/ life insurance. The Age in years is 53. The Other installment plans is none. The Housing is own. The Number of existing credits at this bank is 1. The Job is skilled employee / official. The Number of people being liable to provide maintenance for is 1.0. The Telephone is none. The foreign worker is yes.
\vspace{0.5em}

\textbf{Curated Reasoning by ReSS:}

\vspace{0.5em}
\texttt{<think>}

Step 1: The applicant has no checking account. This absence of a checking account is generally considered a negative indicator for credit risk, as it may suggest limited engagement with formal banking services or less financial transparency.

\vspace{0.5em}

Step 2: The applicant has no other installment plans. Having no other installment plans implies that the applicant is not currently obligated to make additional regular payments elsewhere, which is a positive factor for credit risk as it reduces the likelihood of overextension.

\vspace{0.5em}

Step 3: The applicant’s credit history shows that all existing credits have been paid back duly till now, and there is no record of delayed payments. This is a strong positive indicator, as a history of timely repayments demonstrates reliability and responsible credit behavior.

\vspace{0.5em}

Step 4: The applicant is not classified as female, divorced/separated/married. This condition is neutral in itself but is included as part of the assessment path.

\vspace{0.5em}

Step 5: The applicant has no other debtors or guarantors. This suggests that the applicant is solely responsible for the credit and is not relying on others to support the application, which can be viewed as a positive sign of financial independence.

\vspace{0.5em}

Step 6: The applicant’s savings account or bonds balance is between 500 and less than 1000 DM. This level of savings is a moderate positive indicator, as it shows some financial reserves, though not the highest possible.

\vspace{0.5em}

Step 7: The applicant has one existing credit at this bank, which is less than or equal to 2.5. Having only one existing credit indicates a manageable level of current debt, which is favorable for credit risk.

\vspace{0.5em}

Overall, the following conditions SUPPORT the final classification: the applicant has no other installment plans, has a history of paying back existing credits duly, has no other debtors or guarantors, maintains a moderate level of savings, and has only one existing credit at the bank. The only condition that OPPOSES the classification is the absence of a checking account, which could indicate a higher risk. However, the supporting conditions—especially the strong repayment history and manageable debt—outweigh this single negative factor. These financial conditions collectively indicate that the applicant is classified as "good".
\texttt{</think>}

\texttt{<answer>} good \texttt{</answer>}

\vspace{0.5em}
\end{minipage}
}
\caption{An example of step-by-step reasoning curated by ReSS on Creditg dataset.}
\end{figure}

\newpage
\textbf{Diabetes Dataset}

\begin{figure}[h]
\centering
\fbox{
\begin{minipage}{0.95\linewidth}

\small
\textbf{Original Features:}
\vspace{0.5em}

The Age is 21. The Number of times pregnant is 1. The Diastolic blood pressure is 66. The Triceps skin fold thickness is 23. The Plasma glucose concentration at 2 hours in an oral glucose tolerance test (GTT) is 89. The 2-hour serum insulin is 94. The Body mass index is 28.1. The Diabetes pedigree function is 0.167.
\vspace{0.5em}

\textbf{Curated Reasoning by ReSS:}

\vspace{0.5em}
\texttt{<think>}

Step 1: The patient's plasma glucose concentration at 2 hours in an oral glucose tolerance test is 89, which is less than or equal to the cutoff of 127.500. This value is well within the normal range, indicating that the patient's body is effectively regulating blood glucose levels after a glucose load, and there is no evidence of impaired glucose tolerance or hyperglycemia, both of which are risk factors for diabetes.

\vspace{0.5em}

Step 2: The patient's age is 21, which is less than or equal to the cutoff of 28.500. Younger age is generally associated with a lower risk of type 2 diabetes, as the incidence of diabetes increases with advancing age due to cumulative metabolic stress and potential insulin resistance over time.

\vspace{0.5em}

Step 3: The patient's body mass index (BMI) is 28.1, which is less than or equal to the cutoff of 30.950. While this BMI falls in the overweight category, it is below the threshold commonly associated with a significantly increased risk of type 2 diabetes, which is typically seen at higher BMI values ($\geq 30$)

\vspace{0.5em}

Step 4: The number of times the patient has been pregnant is 1, which is less than or equal to the cutoff of 7.500. A lower number of pregnancies is associated with a reduced risk of gestational diabetes and its contribution to future diabetes risk.

\vspace{0.5em}

Step 5: The patient's diabetes pedigree function is 0.167, which is less than or equal to the cutoff of 0.672. This low value suggests a limited family history or genetic predisposition to diabetes, reducing the overall risk.

\vspace{0.5em}

Overall, all five conditions—normal post-load plasma glucose, young age, BMI below the high-risk threshold, low number of pregnancies, and low diabetes pedigree function—SUPPORT the non-diabetic classification. There are no opposing conditions in this diagnostic path, as each feature value is on the side of reduced diabetes risk. These physiological indicators collectively indicate that the patient is diagnosed as "non-diabetic".

\texttt{</think>}

\texttt{<answer>} non-diabetic \texttt{</answer>}

\vspace{0.5em}
\end{minipage}
}
\caption{An example of step-by-step reasoning curated by ReSS on Diabetes dataset.}
\end{figure}

\newpage

\textbf{Homeloan Dataset}

\begin{figure}[h]
\centering
\fbox{
\begin{minipage}{0.95\linewidth}

\small
\textbf{Original Features:}
\vspace{0.5em}

The gender of the applicant is Male, The marital status of the applicant is Yes, The number of dependents is 2, The education level of the applicant is Graduate, The self-employment status is No, The applicant income is 3073.0, The co-applicant income is 8106.0, The loan amount is 200.0, The loan amount term is 360.0, The credit history value is 1.0, The property area is Urban.
\vspace{0.5em}

\textbf{Curated Reasoning by ReSS:}

\vspace{0.5em}
\texttt{<think>}

Step 1: The applicant’s credit history value is 1.0, which is greater than 0.500. This indicates that the applicant has a positive credit history, suggesting a reliable record of repaying previous loans or credit obligations. This is a strong indicator of creditworthiness and reduces the lender’s risk.

\vspace{0.5em}

Step 2: The property area is Urban, rather than Semiurban. Urban properties are generally considered to have higher market value and better resale potential, which can make the loan less risky for the lender.

\vspace{0.5em}

Step 3: The co-applicant income is 8106.0, which is greater than 612.000. This means the co-applicant brings a substantial additional income, improving the overall household’s repayment capacity and making it more likely that the loan installments can be managed comfortably.

\vspace{0.5em}

Step 4: The loan amount requested is 200.0, which is greater than 166.000. This indicates the applicant is seeking a moderately higher loan amount, which could increase risk if not supported by sufficient income or other positive factors.

\vspace{0.5em}

Step 5: The applicant is married. Being married can sometimes be associated with greater financial stability, as there may be dual incomes or shared financial responsibilities, which can positively influence the ability to repay the loan.

\vspace{0.5em}

Overall, the following conditions support the approval: a strong positive credit history, the property being located in an urban area, a high co-applicant income, and the applicant’s marital status. The only condition that could potentially oppose approval is the moderately higher loan amount requested. However, the applicant’s robust credit history and strong combined income, along with the urban property location, provide substantial support for the ability to manage and repay the loan. These supporting factors clearly outweigh the single opposing factor, leading to a favorable assessment. These characteristics collectively support a "Approved" decision regarding the home loan application.

\texttt{</think>}

\texttt{<answer>} Approved \texttt{</answer>}

\vspace{0.5em}
\end{minipage}
}
\caption{An example of step-by-step reasoning curated by ReSS on Homeloan dataset.}
\end{figure}

\newpage

\textbf{Credit-D Dataset}

\begin{figure}[h]
\centering
\fbox{
\begin{minipage}{0.95\linewidth}

\small
\textbf{Original Features:}
\vspace{0.5em}

The revolving utilization of unsecured lines is 0.704591673. The age of the applicant is 63.0. The number of times the applicant was 30 to 59 days past due without worse delinquency is 0.0. The debt ratio is 0.47144107. The monthly income is 8000.0. The number of open credit lines and loans is 9.0. The number of times the applicant was 90 days late is 0.0. The number of real estate loans or lines is 1.0. The number of times the applicant was 60 to 89 days past due without worse delinquency is 0.0. The number of dependents is 0.0.
\vspace{0.5em}

\textbf{Curated Reasoning by ReSS:}

\vspace{0.5em}
\texttt{<think>}

Step 1: The applicant’s revolving utilization of unsecured lines is 0.704591673, which is greater than 0.435 and less than or equal to 0.773. This means the applicant is using a moderate to high proportion of their available unsecured credit, but not at an extreme level. While higher utilization can signal increased risk, remaining below 0.773 suggests the risk is not excessive.

\vspace{0.5em}

Step 2: The number of times the applicant was 90 days late is 0.0, which is less than or equal to 0.500. This indicates the applicant has not experienced any severe delinquencies, a strong sign of reliable credit behavior and low financial distress risk.

\vspace{0.5em}

Step 3: The number of times the applicant was 30 to 59 days past due without worse delinquency is 0.0, which is less than or equal to 0.500. This further demonstrates the applicant’s consistent on-time payment history, reducing the likelihood of financial distress.

\vspace{0.5em}

Step 4: The number of times the applicant was 60 to 89 days past due without worse delinquency is 0.0, which is less than or equal to 0.500. This reinforces the applicant’s record of avoiding moderate delinquencies, supporting a low risk profile.

\vspace{0.5em}

Step 5: The applicant’s debt ratio is 0.47144107, which is less than or equal to 0.563. This indicates that the applicant’s total monthly debt payments are well-managed relative to their income, further lowering the risk of financial distress.

\vspace{0.5em}

Overall, the conditions that support a low risk of financial distress are the applicant’s absence of any 30-59, 60-89, or 90+ day delinquencies and a manageable debt ratio. The only factor that could suggest elevated risk is the moderately high revolving utilization, but since it remains below the higher threshold of 0.773, this concern is outweighed by the applicant’s strong payment history and controlled debt ratio. These financial conditions collectively indicate that the applicant is classified as "no\_default".

\texttt{</think>}

\texttt{<answer>} no\_default \texttt{</answer>}

\vspace{0.5em}
\end{minipage}
}
\caption{An example of step-by-step reasoning curated by ReSS on Credit-D dataset.}
\end{figure}

\newpage
\newpage

\subsection{Examples of Curated Reasoning Data by Direct Reasoning Curation}
\label{appendix:reasoning_example_DRC}
\textbf{Alzheimer's Disease Dataset}
\begin{figure}[h]
\centering
\fbox{
\begin{minipage}{0.95\linewidth}

\small
\textbf{Original Features:}

The atrophy level of hippocampal is 2. The atrophy level of Amygdala is 2. The atrophy level of entorhinal is 2. The atrophy level of parahippocampal is 2. The atrophy level of medial temporal lobe is 2. The atrophy level of fusiform is 2. The atrophy level of precuneus is 0. The atrophy level of superior parietal is 0. The atrophy level of medial temporal lobe (vs cortex) is 2. The atrophy level of parietal lobe (vs cortex) is 0. The atrophy level of frontal lobe is 2. The atrophy level of temporal lobe is 2. The atrophy level of parietal lobe is 0. The atrophy level of occipital lobe is 0. The overall cortical atrophy is 1. The level of ventricle enlargement is 0. The temporal shape of the lateral ventricle is 2. The frontal shape of the lateral ventricle is 2. The evidence of vascular disease is 0. The age is 88.8 years old. The sex is female. The MMSE score is 19.0. The APOE $\varepsilon4$ status is 1. The marital status is married. The years of education are 13.

\textbf{Curated Reasoning:}

\texttt{<think>}

Step 1: The atrophy level of the hippocampus is 2 (severe atrophy).  \\
Severe hippocampal atrophy is a hallmark of Alzheimer's disease, as this region is critical for memory formation and is typically one of the earliest and most severely affected areas in the disease.

\vspace{0.5em}

Step 2: The atrophy level of the amygdala is 2 (severe atrophy).  \\
Severe amygdala atrophy is commonly observed in Alzheimer's disease and contributes to behavioral and emotional symptoms, further supporting a diagnosis of advanced neurodegeneration.

\vspace{0.5em}

Step 3: The atrophy level of the entorhinal cortex is 2 (severe atrophy).  \\
Severe entorhinal atrophy is highly indicative of Alzheimer's disease, as this region is among the first to show pathological changes and is closely linked to memory impairment.

\vspace{0.5em}

Step 4: The atrophy level of the parahippocampal gyrus is 2 (severe atrophy).  \\
Severe parahippocampal atrophy further supports Alzheimer's disease, as this region is involved in memory encoding and is frequently affected in the disease.

\vspace{0.5em}

Step 5: The atrophy level of the medial temporal lobe is 2 (severe atrophy).  \\
Severe medial temporal lobe atrophy is a strong indicator of Alzheimer's disease, as this area encompasses several structures crucial for memory and is characteristically involved in the disease process.

...

\vspace{0.5em}

Step 13: The APOE $\varepsilon4$ status is 1 (one allele present).  \\
The presence of one APOE $\varepsilon4$ allele increases the risk for Alzheimer's disease, providing genetic support for the diagnosis.

\vspace{0.75em}

Synthesis: \\
- Steps supporting Alzheimer's Disease: Steps 1--8 (severe atrophy in hippocampus, amygdala, entorhinal, parahippocampal, medial temporal, fusiform, frontal, and temporal lobes), Step 9 (pattern of atrophy), Step 10 (ventricle shape changes), Step 11 (MMSE 19.0), Step 12 (advanced age), Step 13 (APOE $\varepsilon4$ carrier). \\
- Steps supporting Mild Cognitive Impairment: Step 11 (MMSE 19.0 could be borderline for MCI, but in context of severe atrophy, supports dementia). \\
- Steps supporting Normal Cognition: None. \\
- Neutral: None.

\vspace{0.5em}

The overwhelming majority of features---especially the pattern of severe atrophy in Alzheimer's-vulnerable regions, moderate cognitive impairment, advanced age, and genetic risk---strongly support Alzheimer's Disease over Mild Cognitive Impairment or Normal Cognition. The evidence is consistent, convergent, and highly specific for Alzheimer's Disease, with no features suggesting normal cognition or only mild impairment.

\vspace{0.5em}

Overall, these characteristics collectively support a ``Alzheimer's Disease'' classification.

\texttt{</think>}

\texttt{<answer>} Alzheimer's Disease \texttt{</answer>}

\vspace{0.5em}
\end{minipage}
}
\caption{An example of step-by-step reasoning obtained by direct reasoning curation on Alzheimer's Disease dataset.}
\end{figure}

\newpage

\textbf{Creditg Dataset}
\begin{figure}[h]
\centering
\fbox{
\begin{minipage}{0.95\linewidth}

\small
\textbf{Original Features:}
\vspace{0.5em}

The Status of existing checking account is $<$ 0 DM. The Duration in month is 6. The Credit history is critical account/ other credits existing (not at this bank). The Purpose is radio/television. The Credit amount is 1169. The Savings account/bonds is unknown/ no savings account. The Present employment since is $\geq 7$. The Installment rate in percentage of disposable income is 4. The Personal status and sex is male : single. The Other debtors / guarantors is none. The Present residence since is 4. The Property is real estate. The Age in years is 67. The Other installment plans is none. The Housing is own. The Number of existing credits at this bank is 2. The Job is skilled employee / official. The Number of people being liable to provide maintenance for is 1.0. The Telephone is yes, registered under the customers name. The foreign worker is yes.
\vspace{0.5em}

\textbf{Curated Reasoning:}

\vspace{0.5em}
\texttt{<think>}

Step 1: The status of the existing checking account is $< 0$ DM. \\
This indicates that the applicant currently has a negative balance or no funds in their checking account, which is generally a negative indicator for creditworthiness as it may suggest liquidity issues or poor cash management.

\vspace{0.5em}

Step 2: The duration in month is 6. \\
A short loan duration reduces the lender's exposure and risk, as the applicant is expected to repay the credit in a relatively short period. This is a positive factor for credit risk.

\vspace{0.5em}

Step 3: The credit history is classified as a critical account or other credits existing (not at this bank). \\
A ``critical account'' status is a strong negative signal, as it suggests past or current difficulties in meeting credit obligations.

\vspace{0.5em}

Step 4: The savings account or bonds status is unknown or indicates no savings account. \\
The absence of savings indicates a lack of financial reserves, which is a negative factor in credit risk assessment.

\vspace{0.5em}

Step 5: The present employment duration is $\geq 7$ years. \\
Long-term employment (at least 7 years) demonstrates stability and a reliable income source, which is a strong positive indicator for creditworthiness.

\vspace{0.5em}

Step 6: The installment rate as a percentage of disposable income is 4. \\
A low installment rate relative to disposable income suggests that the applicant is not overburdened by debt payments, which is a positive sign.

\vspace{0.5em}

...

\vspace{0.5em}

Step 13: The foreign worker status is yes. \\
Being a foreign worker may introduce some uncertainty regarding long-term residence or employment, which is a minor negative factor.

\vspace{0.75em}

Overall, the applicant displays several strong positive indicators, including long-term employment, ownership of real estate and housing, a skilled occupation, a low installment rate, and a short loan duration. These factors support a good credit risk classification. Opposing these are the negative checking account status, critical credit history, lack of savings, and the applicant's status as a foreign worker. However, the stability and asset ownership, combined with manageable debt obligations and employment history, outweigh the negatives. The applicant's overall financial profile supports the classification as ``good''.

\texttt{</think>}

\texttt{<answer>} good \texttt{</answer>}

\vspace{0.5em}
\end{minipage}
}
\caption{An example of step-by-step reasoning obtained by direct reasoning curation on Creditg dataset.}
\end{figure}

\newpage

\textbf{Diabetes Dataset}
\begin{figure}[h]
\centering
\fbox{
\begin{minipage}{0.95\linewidth}

\small
\textbf{Original Features:}
\vspace{0.5em}

The Age is 50. The Number of times pregnant is 6. The Diastolic blood pressure is 72. The Triceps skin fold thickness is 35. The Plasma glucose concentration at 2 hours in an oral glucose tolerance test (GTT) is 148. The 2-hour serum insulin is 0. The Body mass index is 33.6. The Diabetes pedigree function is 0.627.
\vspace{0.5em}

\textbf{Curated Reasoning:}

\vspace{0.5em}
\texttt{<think>}

Step 1: The individual's plasma glucose concentration at 2 hours in an oral glucose tolerance test (GTT) is 148 mg/dL. \\
This value is at the threshold for impaired glucose tolerance (prediabetes) and is suggestive of abnormal glucose metabolism, which is a strong indicator of diabetes risk.

\vspace{0.5em}

Step 2: The body mass index (BMI) is 33.6. \\
This places the individual in the obese category, which is a significant risk factor for the development of type 2 diabetes due to its association with insulin resistance.

\vspace{0.5em}

Step 3: The 2-hour serum insulin is 0. \\
An undetectable insulin level after a glucose challenge is highly abnormal and may indicate severe beta-cell dysfunction or insulin deficiency, both of which are characteristic of diabetes.

\vspace{0.5em}

Step 4: The age is 50 years old. \\
Advancing age is associated with increased risk for type 2 diabetes, as insulin sensitivity tends to decrease and beta-cell function may decline over time.

\vspace{0.5em}

Step 5: The diabetes pedigree function is 0.627. \\
This value suggests a moderate to high familial risk for diabetes, indicating a genetic predisposition.

\vspace{0.5em}

Step 6: The triceps skin fold thickness is 35 mm. \\
This is above average and suggests increased subcutaneous fat, which is often correlated with overall adiposity and increased diabetes risk.

\vspace{0.5em}

Step 7: The diastolic blood pressure is 72 mmHg. \\
This is within the normal range and does not contribute to increased diabetes risk.

\vspace{0.5em}

Step 8: The number of times pregnant is 6. \\
Higher parity has been associated with increased diabetes risk, possibly due to metabolic changes during pregnancy.

\vspace{0.75em}

Overall, the features that support the classification of diabetes are elevated 2-hour plasma glucose (at the diagnostic threshold), obesity (high BMI), undetectable 2-hour insulin (suggesting beta-cell dysfunction), older age, moderate-to-high diabetes pedigree function, increased triceps skin fold thickness, and high parity. The only feature that opposes increased diabetes risk is the normal diastolic blood pressure, which is not a strong protective factor in this context. The multiple strong supporting features, especially the abnormal glucose and insulin findings, clearly outweigh the single opposing condition. Therefore, these clinical and metabolic conditions collectively indicate that the individual is classified as ``diabetic''.

\texttt{</think>}

\texttt{<answer>} diabetic \texttt{</answer>}

\vspace{0.5em}
\end{minipage}
}
\caption{An example of step-by-step reasoning obtained by direct reasoning curation on Diabetes dataset.}
\label{fig:appendix:DRC_example_Diabetes}
\end{figure}

\newpage

\textbf{Homeloan Dataset}
\begin{figure}[h]
\centering
\fbox{
\begin{minipage}{0.95\linewidth}

\small
\textbf{Original Features:}
\vspace{0.5em}

The gender of the applicant is Male, The marital status of the applicant is Yes, The number of dependents is 2, The education level of the applicant is Graduate, The self-employment status is Yes, The applicant income is 5417.0, The co-applicant income is 4196.0, The loan amount is 267.0, The loan amount term is 360.0, The credit history value is 1.0, The property area is Urban, The final\_decision is Approved.

\vspace{0.5em}

\textbf{Curated Reasoning:}

\vspace{0.5em}
\texttt{<think>}

Step 1: The applicant income is 5{,}417.0 and the co-applicant income is 4{,}196.0. \\
This results in a combined monthly income of 9{,}613.0. A higher combined income increases the applicant's ability to afford monthly loan repayments, reducing the risk of default and supporting loan approval.

\vspace{0.5em}

Step 2: The loan amount is 267.0 and the loan term is 360.0 months. \\
This indicates a moderate loan size spread over a standard 30-year term, which generally results in manageable monthly payments relative to the applicant's income, further supporting affordability.

\vspace{0.5em}

Step 3: The credit history value is 1.0. \\
This typically indicates a positive credit history, which is a strong indicator of responsible past borrowing behavior and reduces perceived repayment risk.

\vspace{0.5em}

Step 4: The self-employment status is Yes. \\
While self-employment can sometimes introduce income variability, the relatively high combined income and positive credit history mitigate this risk.

\vspace{0.5em}

Step 5: The number of dependents is 2. \\
While having dependents increases household expenses, the high combined income suggests that the applicant can still comfortably manage both family obligations and loan repayments.

\vspace{0.5em}

Step 6: The education level of the applicant is Graduate. \\
Higher education levels are often associated with better employment prospects and income stability, which supports the applicant's long-term repayment ability.

\vspace{0.75em}

Overall, the supporting conditions include a high combined income, a manageable loan amount and term, a positive credit history, and a graduate-level education. The only potentially opposing condition is the self-employment status, which could introduce income variability; however, this is outweighed by the strong income, positive credit history, and other favorable factors. Therefore, these characteristics collectively support an ``Approved'' decision regarding the home loan application.

\texttt{</think>}

\texttt{<answer>} Approved \texttt{</answer>}

\vspace{0.5em}
\end{minipage}
}
\caption{An example of step-by-step reasoning obtained by direct reasoning curation on Homeloan dataset.}
\end{figure}

\newpage

\textbf{Credit-D Dataset}
\begin{figure}[h]
\centering
\fbox{
\begin{minipage}{0.95\linewidth}

\small
\textbf{Original Features:}
\vspace{0.5em}

The revolving utilization of unsecured lines is 0.704591673. The age of the applicant is 63.0. The number of times the applicant was 30 to 59 days past due without worse delinquency is 0.0. The debt ratio is 0.47144107. The monthly income is 8000.0. The number of open credit lines and loans is 9.0. The number of times the applicant was 90 days late is 0.0. The number of real estate loans or lines is 1.0. The number of times the applicant was 60 to 89 days past due without worse delinquency is 0.0. The number of dependents is 0.0.
\vspace{0.5em}

\textbf{Curated Reasoning:}

\vspace{0.5em}
\texttt{<think>}

Step 1: The revolving utilization of unsecured lines is 0.704591673. This means the individual is using approximately 70\% of their available unsecured credit, which is relatively high and may indicate a greater risk of financial distress due to limited remaining credit capacity

\vspace{0.5em}

Step 2: The age of the applicant is 63.0. Being older may suggest greater financial experience and possibly more stable income sources, which can reduce the risk of financial distress

\vspace{0.5em}

Step 3: The number of times the applicant was 30 to 59 days past due without worse delinquency is 0.0. This indicates no recent short-term delinquencies, which supports financial stability.

\vspace{0.5em}

Step 4: The debt ratio is 0.47144107. This means that the individual's monthly debt payments are about 47\% of their monthly income, which is moderate and generally manageable, suggesting a lower risk of financial distress.

\vspace{0.5em}

Step 5: The monthly income is 8000.0. A relatively high monthly income provides a strong buffer against financial distress, as it increases the ability to meet debt obligations.\

\vspace{0.5em}

Step 6: The number of times the applicant was 90 days late is 0.0, and the number of times the applicant was 60 to 89 days past due without worse delinquency is also 0.0. The absence of any significant delinquencies further supports financial stability.

\vspace{0.75em}

Overall, the high revolving utilization of unsecured lines (Step 1) supports the possibility of financial distress, as it indicates heavy use of available credit. However, several conditions oppose financial distress: the absence of any delinquencies (Steps 3 and 6), a moderate debt ratio (Step 4), high monthly income (Step 5), and the applicant's age (Step 2), which may reflect financial maturity. The opposing conditions—especially the strong payment history and high income—outweigh the supporting condition of high credit utilization. Therefore, these financial conditions collectively indicate that the individual is classified as "no\_default."

\texttt{</think>}

\texttt{<answer>} no\_default \texttt{</answer>}

\vspace{0.5em}
\end{minipage}
}
\caption{An example of step-by-step reasoning obtained by direct reasoning curation on Credit-D dataset.}
\label{fig:appendix:DRC_example_Diabetes}
\end{figure}

\newpage

\subsection{Examples of Delexicalized  Curated  Reasoning}
\label{appendix:reasoning_example_delexicalized_ReSS}
Since delexicalization removes domain-specific semantics, the generated reasoning exhibits minimal variation across datasets. Accordingly, we use examples from the Diabetes dataset as a representative case.
\begin{figure}[h]
\centering
\fbox{
\begin{minipage}{0.95\linewidth}

\small
\textbf{Original Features:}
\vspace{0.5em}

The Feature\_A is 1.0. The Feature\_B is 89.0. The Feature\_C is 66.0. The Feature\_D is 23.0. The Feature\_E is 94.0. The Feature\_F is 28.1. The Feature\_G is 0.2. The Feature\_H is 21.0.

\vspace{0.5em}

\textbf{Curated Reasoning:}

\vspace{0.5em}
\texttt{<think>}

Step 1: Feature\_B is 89.0, which is $\leq$ 127.500.

\vspace{0.5em}

Step 2: Feature\_H is 21.0, which is $\leq$ 28.500.

\vspace{0.5em}

Step 3: Feature\_F is 28.1, which is $\leq$ 30.950.
\vspace{0.5em}

Step 4: Feature\_A is 1.0, which is $\leq$ 7.500.

\vspace{0.5em}

Step 5: Feature\_G is 0.2, which is $\leq$ 0.672.

\vspace{0.5em}

\texttt{</think>}

\texttt{<answer>} Label\_0 \texttt{</answer>}

\vspace{0.5em}
\end{minipage}
}
\caption{An example of delexicalized step-by-step reasoning curated by ReSS on the Diabetes dataset.}

\end{figure}
\section{Hyperparameters and Compute Resources}
\label{appendix:hyperparameters}
\subsection{Decision Tree}
\label{appendix:hyperparameters_DT}
For the Decision Tree baseline, we perform grid search over the following hyperparameter space:

\begin{itemize}
    \item \texttt{max\_depth} $\in \{4, 5, 6, 7\}$
    \item \texttt{min\_samples\_split} $\in \{2, 5, 10, 20\}$
    \item \texttt{min\_samples\_leaf} $\in \{1, 2, 5, 10\}$
    \item \texttt{criterion} $\in \{\text{gini}, \text{entropy}\}$
\end{itemize}

The optimal hyperparameters are selected based on validation accuracy.

\subsection{XGBoost}

For the XGBoost baseline, we perform grid search over the following hyperparameter space:

\begin{itemize}
    \item \texttt{max\_depth} $\in \{2, 4, 6, 8, 10, 12\}$
    \item \texttt{alpha} $\in \{10^{-8}, 10^{-7}, 10^{-6}, 10^{-5}, 10^{-4}, 10^{-3}, 10^{-2}, 10^{-1}, 1\}$
    \item \texttt{lambda} $\in \{10^{-8}, 10^{-7}, 10^{-6}, 10^{-5}, 10^{-4}, 10^{-3}, 10^{-2}, 10^{-1}, 1\}$
    \item \texttt{eta} $\in \{0.01, 0.03, 0.1, 0.3\}$
\end{itemize}

The optimal hyperparameters are selected based on validation accuracy.

\subsection{TabNet}

For the TabNet baseline, we perform grid search over the following hyperparameter space:

\begin{itemize}
    \item \texttt{n\_d} $\in \{8, 16\}$
    \item \texttt{n\_a} $\in \{8, 16\}$
    \item \texttt{n\_steps} $\in \{3, 5\}$
    \item \texttt{gamma} $\in \{1.0, 1.3\}$
    \item \texttt{lambda\_sparse} $\in \{10^{-3}\}$
    \item \texttt{mask\_type} $\in \{\texttt{sparsemax}, \texttt{entmax}\}$
    \item \texttt{learning\_rate} $\in \{0.02, 0.01\}$
\end{itemize}
The optimal hyperparameters are selected based on validation accuracy.

\subsection{Supervised Fine-Tuning}
\label{appendix:hyperparameters_SFT}
For SFT, we adopt the trl~\cite{vonwerra2022trl} framework and train the models using the AdamW optimizer with a learning rate of $1 \times 10^{-5}$, weight decay set to 0, and a cosine learning-rate scheduler with a warmup ratio of $0.03$.
We perform full-parameter fine-tuning with a training batch size of 2. The model is trained for 20 epochs on the HomeLoan, Creditg, and Diabetes datasets, while only three epochs are used for the AD and Credit-D datasets. Model evaluation is performed every 50 training steps.
During inference, we use greedy decoding with temperature set to 0.
\subsection{Reinforcement Learning}
\label{appendix:hyperparameters_RL}
For Direct RL, we use the DisCO algorithm~\cite{li2025disco} due to its observed better performance than GRPO. We train the model with a batch size of 128 samples, a mini-batch size of 64,  maximum response length of 1024. For each prompt, we generate 8 responses and use only accuracy based rewards. Optimization is performed with AdamW optimizer with a constant learning rate of $2e^{-6}$ and a weight decay of 0.01. The temperature is set to 0.7 for training and 0.0 for evaluation. Models are trained for 320 RL steps, with evaluations conducted every 10 steps.

For ReSS+RL, we train the model for 200 RL steps with a temperature of 1.0, as the starting model has already undergone supervised fine-tuning.

\subsection{Compute Resources}
\label{appendix:compute}
All supervised fine-tuning and reinforcement learning experiments are conducted on a single node with 4 NVIDIA A100 GPUs (80GB each). For supervised fine-tuning, a single run typically takes approximately 8 to 24 hours depending on the dataset size. For reinforcement learning, a single run takes around 8 hours. Reasoning data curation is performed via the GPT-4.1 API.

\newpage
\section{Supplementary Experimental Results}
\label{appendix:supplementary_results}

\subsection{Explainability without Symbolic Guidance}
\label{appendix:reasoning_explainability}
We further analyze the explainability of the generated reasoning by comparing ReSS,
trained without augmented reasoning data, with a DRC+SFT baseline,
whose reasoning data is generated without symbolic scaffold constraints.
This analysis focuses on explanation sufficiency and necessity,
using the same feature-masking protocol as in the main experiments.

\begin{figure}[h]
  \centering
  \includegraphics[width=0.95\columnwidth]{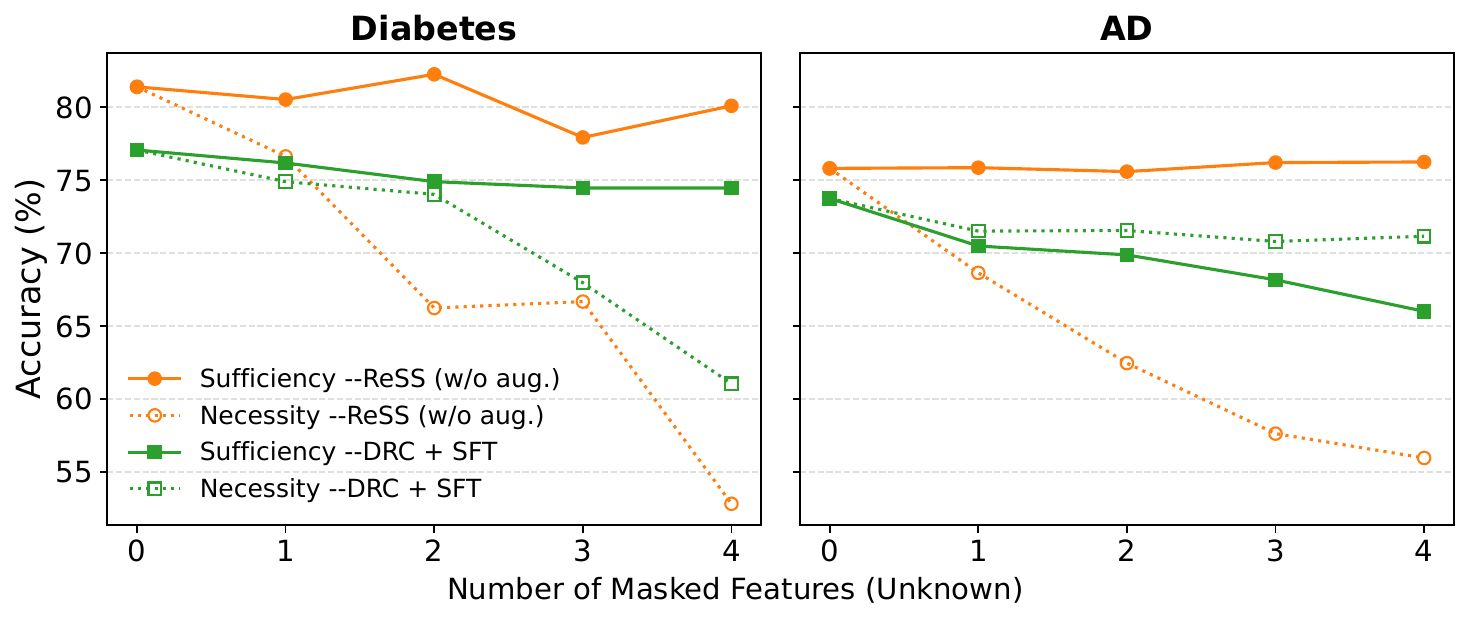}
  \caption{Sufficiency and necessity analysis via feature masking on Diabetes and AD. We compare the ReSS model trained without augmented reasoning data against the DRC+SFT baseline, whose reasoning data is generated without symbolic scaffold constraints.}
  \label{fig:DRC_ReSS_masking}
\end{figure}

Figure~\ref{fig:DRC_ReSS_masking} shows the sufficiency and necessity curves
on Diabetes and AD.
For ReSS, masking unused features results in only minor accuracy changes,
while masking explanation-referenced features leads to a sharp and monotonic
performance drop, indicating strong explanation necessity.
In contrast, DRC+SFT consistently exhibits substantially weaker necessity.
On Diabetes, masking features referenced by the explanation causes a much smaller
accuracy degradation compared to ReSS.
The issue becomes more pronounced on AD, where the accuracy drop under necessity
masking is minimal and, in some cases, even smaller than that observed when masking
unused features, revealing a clear violation of explanation necessity.

This behavior can be attributed to the absence of symbolic guidance during
reasoning data generation.
Without symbolic scaffold constraints, the LLM tends to incorporate a large number
of input features into the explanation, rather than selectively identifying
decision-critical evidence.
This over-inclusive behavior is particularly detrimental for high-dimensional
datasets such as AD, which contains 25 input features.
Under DRC+SFT, the generated explanations reference an average of 14 features,
many of which are redundant or only weakly correlated with the prediction outcome.

From an explainability perspective, such reasoning fails to distinguish truly
influential features from incidental information, leading to substantially
degraded explanation necessity.
In contrast, the symbolic scaffolds in ReSS explicitly encourage structured
reasoning, ensuring that the features highlighted by the explanation are genuinely
relevant to the model’s decision.

\subsection{Knowledge helps our ReSS-trained LLM to correct mistakes of Decision tree}
\label{appendix:cherry_pick}

Decision trees are purely data-driven models. As a result, some predictions may reflect dataset-specific artifacts or spurious correlations rather than applying domain-plausible rules. While this can introduce a small fraction of noisy or counterintuitive reasoning examples  for training, the majority of decision-tree-derived reasoning paths remain consistent with domain commonsense. When trained on such data, the language model does not simply memorize individual paths, but instead learns implicit explanatory structures and domain-level regularities underlying the paths. Consequently, during inference, the model may generate predictions that deviate from specific decision-tree leaf outcomes while remaining aligned with learned domain-level regularities.

To illustrate this phenomenon, we present a qualitative case study on a testing data from the Diabetes dataset shown in Figure~\ref{fig:DT_vs_ReSS_example}. Along this decision path, the decision tree assigns a \textit{non-diabetic} label, which reflects the empirical training label distribution observed in this localized region of the feature space but is wrong. However,  every condition along the path corresponds to a well-established risk factor for diabetes according to the domain knowledge. {In contrast}, our fine-tuned LLM generates a rationale that faithfully follows the same decision path conditions, explicitly referencing each split and providing medically plausible interpretations for them. The generated reasoning emphasizes the cumulative risk implied by high elevated 2-hour plasma glucose levels, extreme obesity, elevated diabetes pedigree function, and high blood pressure, and finally predicts the correct {\it diabetic} label.  


\begin{figure}[!h]
\centering
\fbox{
\begin{minipage}{0.98\linewidth}
\small
\textbf{Original Features:}

The Age is 23. The Number of times pregnant is 0. The Diastolic blood pressure is 90. The Triceps skin fold thickness is 27. The Plasma glucose concentration at 2 hours in an oral glucose tolerance test (GTT) is 179. The 2-hour serum insulin is 0. The Body mass index is 44.1. The Diabetes pedigree function is 0.686.
\vspace{0.2em}

\textbf{Decision Tree Path:}

Glucose $>$ 127.500,

BMI $>$ 29.950,

DiabetesPedigreeFunction $>$ 0.434,

BMI $>$ 40.100,

BloodPressure $>$ 80.000,

Answer: non-diabetic

\vspace{0.2em}

\textbf{Generated Reasoning by a ReSS-trained model:}

\texttt{<think>}

Step 1: The patient's plasma glucose concentration at 2 hours in an oral glucose tolerance test is 179, which is greater than 127.500. This elevated glucose level indicates impaired glucose tolerance and suggests that the patient’s body is not effectively clearing glucose from the bloodstream, a hallmark of diabetes mellitus.

\vspace{0.2em}

Step 2: The body mass index (BMI) is 44.1, which is greater than 29.950. A BMI in this range is classified as severe obesity, which is a strong risk factor for insulin resistance and the development of type 2 diabetes due to increased adiposity and associated metabolic disturbances.\

\vspace{0.2em}

Step 3: The diabetes pedigree function is 0.686, which is greater than 0.434. This value reflects a higher familial predisposition to diabetes, indicating that genetic factors may contribute significantly to the patient’s risk profile.

\vspace{0.2em}

Step 4: The BMI is again considered and is 44.1, which is greater than 40.100. This further emphasizes the presence of extreme obesity, which is associated with a markedly increased risk of developing diabetes due to profound insulin resistance and chronic low-grade inflammation.

\vspace{0.2em}
Step 5: The diastolic blood pressure is 90, which is greater than 80.000. Elevated blood pressure is a component of metabolic syndrome and is commonly associated with insulin resistance and increased risk of diabetes.

\vspace{0.2em}

Overall, the following conditions SUPPORT the final classification: the markedly elevated plasma glucose concentration, the extremely high BMI (noted twice), the elevated diabetes pedigree function, and the persistently high diastolic blood pressure. Each of these factors is well-established in the literature as a significant risk or diagnostic indicator for diabetes. There are no conditions in this path that OPPOSE the classification; all features point toward increased diabetes risk. The convergence of severe obesity, hyperglycemia, and abnormal blood pressure strongly outweighs any opposing considerations. These physiological indicators collectively  indicate that the patient is diagnosed as "diabetic".

\texttt{<think>} 

\texttt{<answer>} diabetic
\texttt{<answer>}

\vspace{0.2em}
\end{minipage}
}
\caption{An illustrative testing example from the Diabetes dataset.}

\label{fig:DT_vs_ReSS_example}
\end{figure}

\subsection{Ablation Studies on Conducting RL after ReSS }\label{app:SFTRL}
\begin{figure}[h]
  \centering
   \begin{minipage}[t]{0.48\columnwidth}
    \centering
    \includegraphics[width=\linewidth]{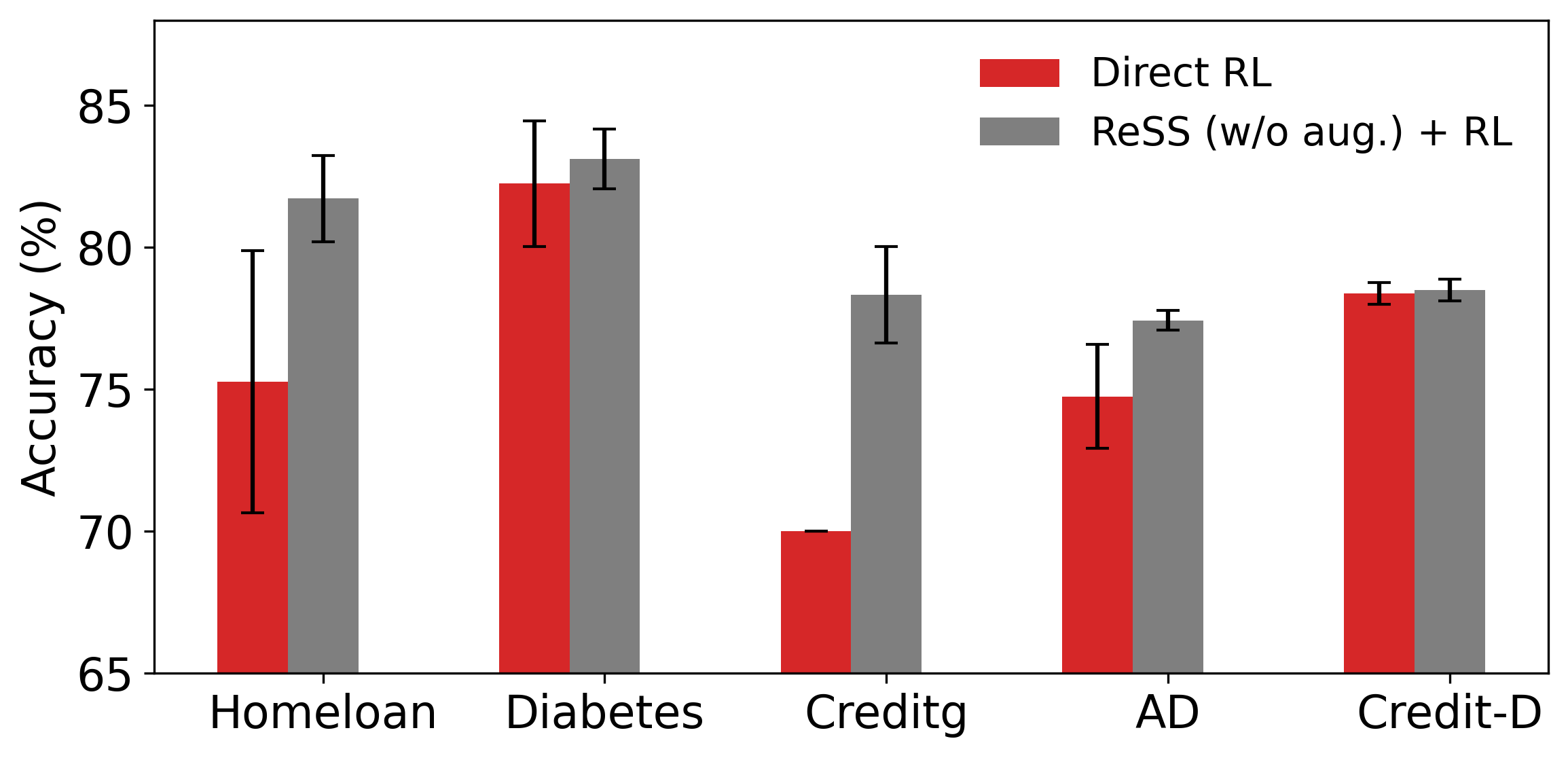}
    \caption{Direct RL vs. ReSS (w/o aug.) + RL.}

    \label{fig:ReSS_RL_Direct_RL}
  \end{minipage}
  \hfill
 \begin{minipage}[t]{0.48\columnwidth}
    \centering
    \includegraphics[width=\linewidth]{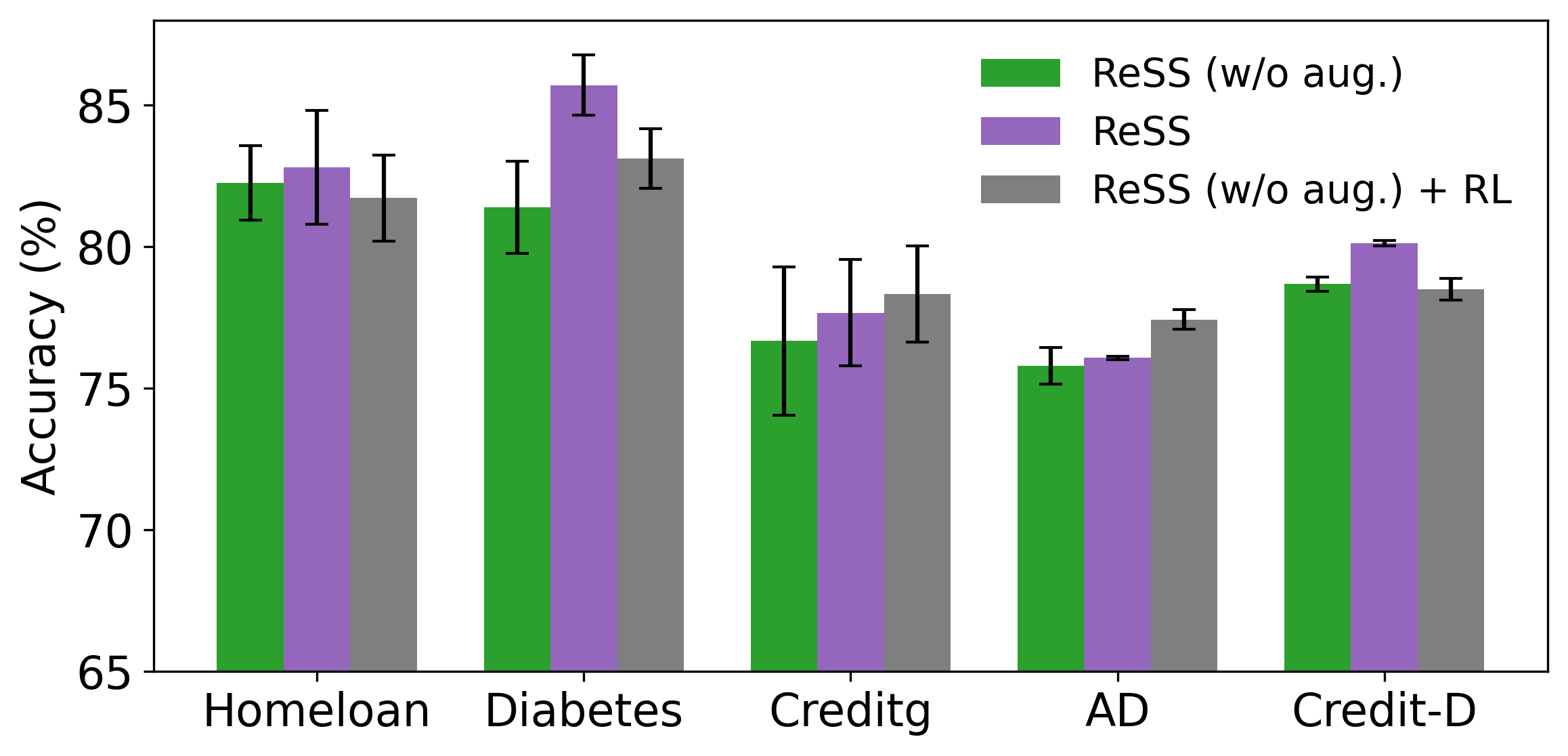}
    \caption{ReSS vs. ReSS (w/o aug.) + RL.}

    \label{fig:RL_aug_Ress}
  \end{minipage}
\end{figure}

The results are shown in Figure~\ref{fig:ReSS_RL_Direct_RL} and Figure~\ref{fig:RL_aug_Ress}. 

\newpage

\section*{NeurIPS Paper Checklist}


\begin{enumerate}

\item {\bf Claims}
    \item[] Question: Do the main claims made in the abstract and introduction accurately reflect the paper's contributions and scope?
    \item[] Answer: \answerYes{}
    \item[] Justification: The abstract and introduction (Section~\ref{sec:intro}) clearly state our contributions: the ReSS framework using decision-tree paths as symbolic scaffolds, the scaffold-invariant data augmentation strategy, the faithfulness and explainability metrics, and the experimental validation on five tabular datasets.


\item {\bf Limitations}
    \item[] Question: Does the paper discuss the limitations of the work performed by the authors?
    \item[] Answer: \answerYes{}
    \item[] Justification: We discuss limitations in the Conclusion, noting that ReSS is validated in financial and medical domains where LLMs possess strong intrinsic domain knowledge, and that its effectiveness in other domains remains future work.


\item {\bf Theory assumptions and proofs}
    \item[] Question: For each theoretical result, does the paper provide the full set of assumptions and a complete (and correct) proof?
    \item[] Answer: \answerNA{}
    \item[] Justification: This paper does not present formal theoretical results; the contributions are methodological and empirical.


    \item {\bf Experimental result reproducibility}
    \item[] Question: Does the paper fully disclose all the information needed to reproduce the main experimental results of the paper to the extent that it affects the main claims and/or conclusions of the paper (regardless of whether the code and data are provided or not)?
    \item[] Answer: \answerYes{}
    \item[] Justification: We describe the full pipeline in Section~\ref{sec:method}, including decision-tree training, scaffold extraction, reasoning data curation, scaffold-invariant augmentation, and SFT. All prompts, serialization templates, and hyperparameter settings are provided in Appendix.


\item {\bf Open access to data and code}
    \item[] Question: Does the paper provide open access to the data and code, with sufficient instructions to faithfully reproduce the main experimental results, as described in supplemental material?
    \item[] Answer: \answerNo{}
    \item[] Justification: Code will be released upon acceptance.


\item {\bf Experimental setting/details}
    \item[] Question: Does the paper specify all the training and test details (e.g., data splits, hyperparameters, how they were chosen, type of optimizer) necessary to understand the results?
    \item[] Answer: \answerYes{}
    \item[] Justification: Data splits are specified in Table~\ref{tab:combined} and Section~\ref{sec:experiments}. All prompts, serialization templates, and hyperparameter settings are provided in Appendix.


\item {\bf Experiment statistical significance}
    \item[] Question: Does the paper report error bars suitably and correctly defined or other appropriate information about the statistical significance of the experiments?
    \item[] Answer: \answerYes{}
    \item[] Justification: All results are averaged over three random seeds with standard deviations.


\item {\bf Experiments compute resources}
    \item[] Question: For each experiment, does the paper provide sufficient information on the computer resources (type of compute workers, memory, time of execution) needed to reproduce the experiments?
    \item[] Answer: \answerYes{} 
    \item[] Justification: Details can be found at Appendix~\ref{appendix:compute}
    
\item {\bf Code of ethics}
    \item[] Question: Does the research conducted in the paper conform, in every respect, with the NeurIPS Code of Ethics \url{https://neurips.cc/public/EthicsGuidelines}?
    \item[] Answer: \answerYes{}
    \item[] Justification: The research conforms with the NeurIPS Code of Ethics. 

\item {\bf Broader impacts}
    \item[] Question: Does the paper discuss both potential positive societal impacts and negative societal impacts of the work performed?
    \item[] Answer: \answerNA{}
    \item[] Justification: This paper is methodological research on training faithful reasoning models for tabular prediction, and is not tied to any particular deployed application.
    
\item {\bf Safeguards}
    \item[] Question: Does the paper describe safeguards that have been put in place for responsible release of data or models that have a high risk for misuse (e.g., pre-trained language models, image generators, or scraped datasets)?
    \item[] Answer: \answerNA{}
    \item[] Justification: This paper does not release any pretrained generative models, scraped datasets, or other assets with a high risk of misuse. The fine-tuned models are narrow tabular classifiers trained on publicly available benchmarks, and pose no foreseeable dual-use concerns.

\item {\bf Licenses for existing assets}
    \item[] Question: Are the creators or original owners of assets (e.g., code, data, models), used in the paper, properly credited and are the license and terms of use explicitly mentioned and properly respected?
    \item[] Answer: \answerYes{}
    \item[] Justification: All datasets and baselines are properly cited, and their licenses and terms of use are respected.


\item {\bf New assets}
    \item[] Question: Are new assets introduced in the paper well documented and is the documentation provided alongside the assets?
    \item[] Answer: \answerNA{}
    \item[] Justification: We do not release new datasets or pretrained models with this submission. Curated reasoning data is described in detail in the appendix to support reproduction.


\item {\bf Crowdsourcing and research with human subjects}
    \item[] Question: For crowdsourcing experiments and research with human subjects, does the paper include the full text of instructions given to participants and screenshots, if applicable, as well as details about compensation (if any)? 
    \item[] Answer: \answerNA{}
    \item[] Justification: This research does not involve crowdsourcing or human subjects.

\item {\bf Institutional review board (IRB) approvals or equivalent for research with human subjects}
    \item[] Question: Does the paper describe potential risks incurred by study participants, whether such risks were disclosed to the subjects, and whether Institutional Review Board (IRB) approvals (or an equivalent approval/review based on the requirements of your country or institution) were obtained?
    \item[] Answer: \answerNA{}
    \item[] Justification: This research does not involve human subjects, and all datasets used are publicly available, de-identified benchmarks.

\item {\bf Declaration of LLM usage}
    \item[] Question: Does the paper describe the usage of LLMs if it is an important, original, or non-standard component of the core methods in this research? Note that if the LLM is used only for writing, editing, or formatting purposes and does \emph{not} impact the core methodology, scientific rigor, or originality of the research, declaration is not required.
    \item[] Answer: \answerYes{}
    \item[] Justification: LLMs are a core component of our methodology. GPT-4.1 is used to generate the symbolic-scaffold-grounded reasoning data, and Qwen-2.5-3B-Instruct is fine-tuned as the target reasoning model.

\end{enumerate}
\end{document}